\documentclass[lettersize,journal]{IEEEtran}
\usepackage{amsmath,amsfonts}
\usepackage{algorithmic}
\usepackage{array}
\usepackage[caption=false,font=normalsize,labelfont=sf,textfont=sf]{subfig}
\usepackage{textcomp}
\usepackage{stfloats}
\usepackage{url}
\usepackage{verbatim}
\usepackage{graphicx}
\usepackage{cite}
\usepackage{topcapt}

\usepackage{graphics}
\usepackage{epsfig}

\usepackage{cite}
\usepackage[dvipsnames]{xcolor}
\usepackage{xurl}
\usepackage{hyperref}
\hypersetup{
    colorlinks,
    linkcolor={red!50!black},
    citecolor={blue!50!black},
    urlcolor={blue!80!black}
}
\usepackage{bm}
\usepackage{breqn}
\usepackage{amssymb}
\usepackage{tabularx}
\usepackage{booktabs}
\usepackage{dcolumn}
\newcolumntype{d}[1]{D..{#1}}
\newcolumntype{C}{>{\centering\arraybackslash}X}
\usepackage{soul}
\usepackage[linesnumbered,ruled,vlined]{algorithm2e} 
\usepackage{setspace}

\usepackage{amsthm}

\newcommand{\gatekeeper}{\texttt{gatekeeper}}

\usepackage{acronym}
\acrodef{SDF}{Signed Distance Field}
\acrodef{ESDF}{Euclidean Signed Distance Field}
\acrodef{SFC}{Safe Flight Corridor}
\acrodef{DMP}{Distance Map Planner}
\acrodef{MPC}{Model Predictive Control}
\acrodef{CBF}{Control Barrier Function}
\acrodef{RoS}[RoS]{Rate of Spread}

\newcommand{\reals}{\mathbb{R}}
\newcommand{\R}{\reals}
\newcommand{\Rnonneg}{\reals_{\geq 0}}
\newcommand{\Rplus}{\reals_{>0}}

\newcommand{\naturals}{\mathbb{N}}

\newcommand{\ball}{\mathbb{B}}

\newcommand{\Bcal}{\mathcal{B}}
\newcommand{\Ccal}{\mathcal{C}}

\newcommand{\Ical}{\mathcal{I}}
\newcommand{\Ncal}{\mathcal{N}}

\newcommand{\Scal}{\mathcal{S}}
\newcommand{\Tcal}{\mathcal{T}}
\newcommand{\Ucal}{\mathcal{U}}
\newcommand{\Xcal}{\mathcal{X}}

\newcommand{\classK}{class~$\mathcal{K}$ }

\newcommand{\classKL}{class~$\mathcal{KL}$ }

\newcommand{\eqn}[1]{\begin{align} #1 \end{align}}
\newcommand{\eqnN}[1]{\begin{align*} #1 \end{align*}}

\newcommand{\bmat}[1]{\begin{bmatrix}#1\end{bmatrix}}

\newcommand{\norm}[1]{\left\Vert #1 \right \Vert}

\newcommand{\argmin}[1]{\underset{#1}{\text{argmin}}}

\newenvironment{nospaceflalign*}
 {\setlength{\abovedisplayskip}{0pt}\setlength{\belowdisplayskip}{0pt}%
  \csname flalign*\endcsname}
 {\csname endflalign*\endcsname\ignorespacesafterend}

\theoremstyle{plain}
\newtheorem{theorem}{Theorem}

\newtheorem{definition}{Definition}

\theoremstyle{definition}

\newtheorem{assumption}{Assumption}
\newtheorem{problem}{Problem}
\newtheorem{remark}{Remark}

\theoremstyle{remark}

\usepackage{etoolbox}
\makeatletter
\patchcmd{\@makecaption}
  {\scshape}
  {}
  {}
  {}
\makeatletter
\patchcmd{\@makecaption}
  {\\}
  {.\ }
  {}
  {}
\makeatother

\usepackage[modulo, switch]{lineno}

\usepackage{fancyhdr}
\fancypagestyle{FirstPage}{
\chead{Accepted for publication at IEEE Transactions on Robotics (T-RO) 2024. \\ \small{© 2024 IEEE.  Personal use of this material is permitted.  Permission from IEEE must be obtained for all other uses, in any current or future media, including reprinting/republishing this material for advertising or promotional purposes, creating new collective works, for resale or redistribution to servers or lists, or reuse of any copyrighted component of this work in other works.}}
}

\begin{document}
\title{\gatekeeper{}: Online Safety Verification and Control for Nonlinear Systems in Dynamic Environments
}

\author{
Devansh Ramgopal Agrawal, Ruichang Chen and Dimitra Panagou
\thanks{The authors would like to acknowledge the support of the National
Science Foundation (NSF) under grant no. 1942907.}
\thanks{Devansh R Agrawal is with the Department of Aerospace Engineering, University of Michigan; Ruichang Chen is with the Department of Electrical and Computer Engineering; Dimitra Panagou is with the Department of Robotics and the Department of Aerospace Engineering, University of Michigan, Ann Arbor, USA. {\tt\small \{devansh, chenrc, dpanagou\}@umich.edu}}%
}

\maketitle
\thispagestyle{empty}
\pagestyle{empty}

\begin{abstract}
This paper presents the \gatekeeper{} algorithm, a real-time and computationally-lightweight method that ensures that trajectories of a nonlinear system satisfy safety constraints despite sensing limitations.  \gatekeeper{} integrates with existing path planners and feedback controllers by introducing an additional verification step to ensure that proposed trajectories can be executed safely, despite nonlinear dynamics subject to bounded disturbances, input constraints and partial knowledge of the environment. Our key contribution is that (A) we propose an algorithm to recursively construct safe trajectories by numerically forward propagating the system over a (short) finite horizon, and (B) we prove that tracking such a trajectory ensures the system remains safe for all future time, i.e., beyond the finite horizon. We demonstrate the method in a simulation of a dynamic firefighting mission, and in physical experiments of a quadrotor navigating in an obstacle environment that is sensed online. We also provide comparisons against the state-of-the-art techniques for similar problems.
\end{abstract}

\thispagestyle{FirstPage}


Code and videos are available here: \cite{gatekeeperRepo}.

\section{INTRODUCTION}
\label{sec:introduction}

Designing autonomous systems with strict guarantees of safety is still a  bottleneck to deploying such systems in the real world. Safety is often posed as requiring the system's trajectories to lie within a set of allowable states, called the safe set. In this paper, we consider the case where the safe set is not known a priori, but is rather built on-the-fly via the system outputs (sensor measurements). More specifically,  we consider the problem where a robot with limited sensing capabilities (hence limited information about the environment) has to move while remaining safe under some mild assumptions on the evolution of the environment, to be stated in detail below.

Navigating within a non-convex safe set is often tackled by path planning techniques~\cite{lavalle2006planning, karaman2011anytime, webb2013kinodynamic, richter2016polynomial}. 
Typically a planner generates reference (or nominal) trajectories based on a simplified (e.g., linearized or kinematic) model of the system. However, the reference trajectories may not be trackable by the actual nonlinear system dynamics, and as a result safety constraints may be violated. Furthermore, when trajectories are planned over finite horizons, without recursive feasibility guarantees a planner may fail to find a trajectories, leading to safety violations. This is particularly relevant and challenging when operating in dynamic environments. 

In this paper, we propose a technique to bridge path planners (that can solve the nonconvex trajectory generation problem) and controllers (that have robust stability guarantees) in a way that ensures safety.  \gatekeeper{} takes inspiration from~\cite{tordesillas2021faster} and~\cite{singletary2022onboard}, both of which also employ the idea of a backup planner/controller. Conceptually, a backup controller is a feedback controller that drives the system to a set of states that are safe (referred to as the backup safe set), and keeps the system in this set. For example, for a quadrotor navigating in an environment with static obstacles, a backup controller could be one that causes the quadrotor to hover in place.

In \gatekeeper{}, the idea is that given a nominal trajectory generated by the path planner (potentially unsafe and/or not dynamically feasible) we construct a  ``committed trajectory" using a backup controller. To do this, at each iteration of \gatekeeper{}, we simulate a controller that tracks the nominal trajectory upto some switching time, and executes the backup controller thereafter. The trajectory with the largest switching time that is valid (as defined in Def~\ref{def:robustly_valid}) becomes the committed trajectory. Thus, each committed trajectory is, by construction, guaranteed to be defined, feasible, and safe for all future time. The controller always tracks the last committed trajectory, thereby ensuring safety. This paper's key contribution is the algorithm to construct such committed trajectories, and a proof that the proposed approach ensures the closed-loop system remains safe. Furthermore, we explicitly account for robustness against disturbances and state-estimation error since naive approaches to robustification can lead to undesired deadlock. The overall algorithm is computationally efficient compared to similar methods, e.g. \ac{MPC}. In our simulations~\ref{sec:sims_and_exps}, \gatekeeper{} was approximately 3-10 times faster than \ac{MPC}. \gatekeeper{}'s primary limitation is that there must exist a backup controller and set. Some robotic systems and environments may not admit these components. Our focus in this paper is on systems where one can find a suitable backup controller and set, and demonstrate how this can be employed to ensure safety.

In summary, this work has the following contributions:
\begin{itemize}
\item A framework to bridge path planners with tracking controllers in order to convert nominal/desired trajectories (generated by the path planner) into committed trajectories that the tracking controller can track safely. 
\item A formal proof that the robotic system will remain safe for all future time under the stated assumptions.
\end{itemize}

In particular, the new contributions of this version with respect to the conference paper~\cite{gatekeeper_iros} are:
\begin{itemize}
\item Theoretical: A robustification of the verification conditions in~\cite{gatekeeper_iros} to also account for state estimation errors. We have also simplified the verification conditions. 
\item Experimental: A demonstration of the algorithm applied to quadrotors flying through an unknown environment, constructing a map of the environment online, and filtering human pilot commands to ensure collision avoidance. 
\end{itemize}

A worked analytic example is provided in the appendix, to help illustrate the key concepts of the paper.

\subsubsection*{Paper Organization} 
In section~\ref{sec:related_work} we review a few of the leading paradigms for safety-critical path planning and control. In section~\ref{sec:motivation}, we describe the key idea underpinning  \gatekeeper{}. In sections~\ref{sec:problem}, \ref{sec:solution} we formally define the problem and describe our proposed solution. Finally, in section~\ref{sec:sims_and_exps} simulations and experiments are used to demonstrate the method, and specific implementation details are discussed. 

\subsubsection*{Notation} 
Let $\naturals = \{0, 1, 2, ... \}$, and $\R, \Rplus, \Rnonneg$ denote the set of reals, positive reals, and non-negative reals. Lowercase $t \in \R$ is used for specific time points, while uppercase $T \in \R$ is for durations. $\norm{\cdot}$ refers to the vector 2-norm. Closed balls are denoted $\ball(r) = \{ x : \norm{x} \leq r\}$. For sets $A,B$, $A \subset B$ means $x \in A \implies x \in B$, and $A \ominus B$, $A \oplus B$ are the Pontryagin set difference and the Minkowski sum of sets $A, B$. A function $\alpha: \Rnonneg \to \Rnonneg$ is \classK if it is continuous, strictly increasing and $\alpha(0) = 0$. $\beta: \Rnonneg \times \Rnonneg \to \Rnonneg$ is a \classKL function if it is continuous, for each $t \geq 0$, $\beta(\cdot, t)$ is class~$\mathcal{K}$, and for each $r > 0$, $\beta(r, \cdot)$ is strictly decreasing and $\lim_{t \to \infty} \beta(r, t) = 0$. See also Table~\ref{tab:nomenclature}.

\section{RELATED WORK}
\label{sec:related_work}

\begin{figure*}
     \centering
     \includegraphics[width=0.98\linewidth]{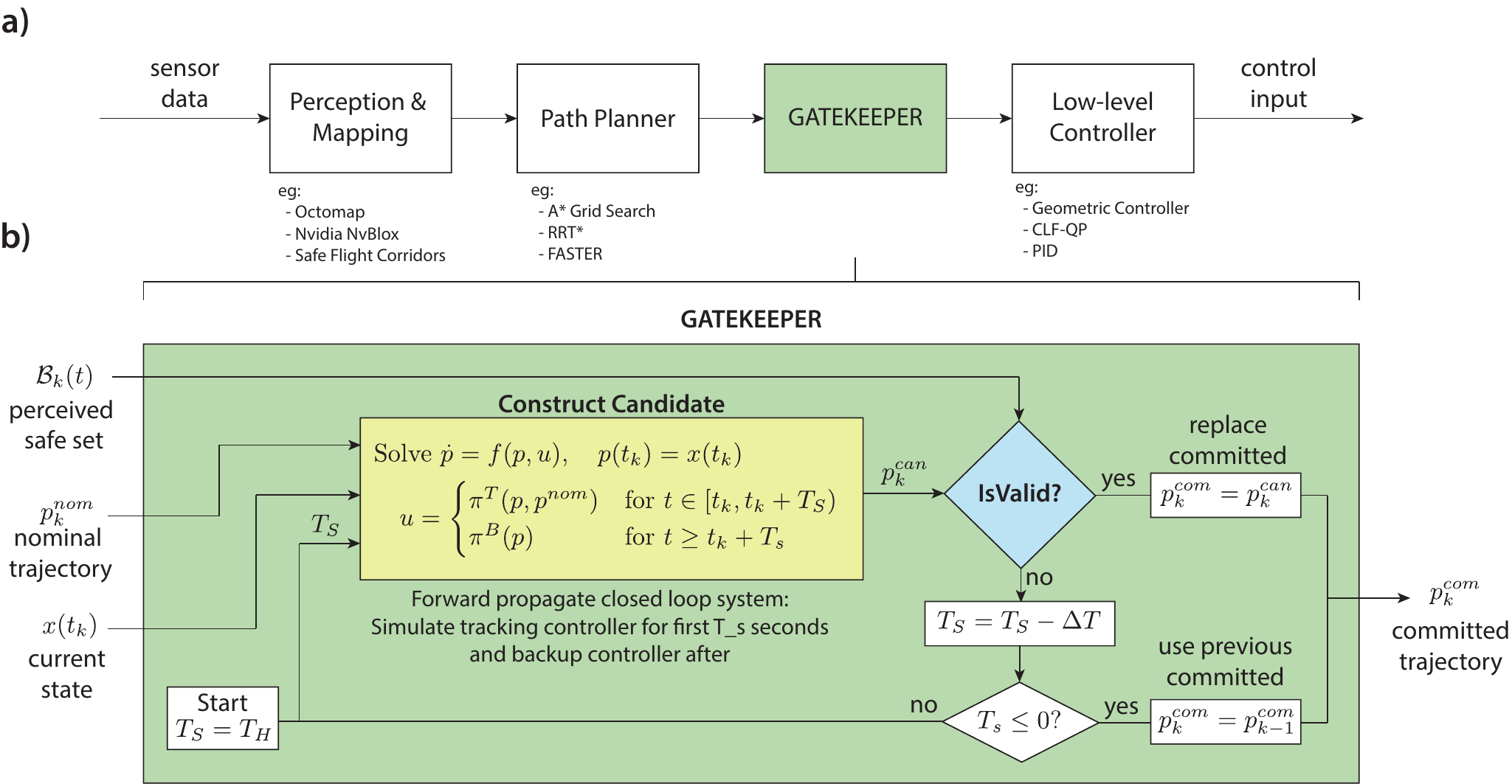}
     \caption{Block Diagram describing the \gatekeeper{} algorithm. (a) shows that \gatekeeper{} is an additional module that fits within the common perception-planning-control stack of a robotic system. (b) is a pictorial representation of Algorithm~\ref{alg:full}. }
     \label{fig:block_diagram}
 \end{figure*}

A wide range of architectures and approaches have been proposed to tackle safety-critical planning and control, especially when the environment is sensed online. A generic perception planning and control stack is depicted in Fig.~\ref{fig:block_diagram}a.

One approach is to encode the safety constraints in the path-planning module. In this case, the world is  represented using a grid-world, or through simplified geometric primitives like obstacle points, or planes to depict the walls. From this representation, a path is generated to avoid obstacles using, for instance, grid-search techniques~\cite{harabor2011online} or sampling~\cite{karaman2011anytime}. These paths can then be modified to avoid the obstacles, e.g.~\cite{liu2017planning}. However since the path was generated without considering the closed-loop behavior of the nonlinear dynamics of the system and the controller, the robot may not execute the planned path exactly. Therefore, safety may not be guaranteed. 

A second approach is to encode the safety constraints at the controller. In recent years, methods based on \acp{CBF}~\cite{ames2019control} have been developed to ensure that a system remains within a specified safe set while tracking a desired control input. These methods however require the safe set to be known apriori, represented by a scalar function $h : \Xcal \to \R$ that is continuously differentiable, and satisfies an invariance condition (see for e.g. Def. 2 of~\cite{ames2019control}). For certain classes of systems and safe sets, constructive methods exist to design $h$, but these do not handle time-varying or multiple safety conditions well~~\cite{cortez2020correct, breeden2021guaranteed, abate2020enforcing, llanes2021safety, agrawal2021constructive}. For specific system models, it is sometimes possible to construct suitable planners and controllers, e.g.~\cite{chen2021fastrack, bajcsy2019efficient}. Alternatively, offline and computationally expensive methods based on Hamilton-Jacobi reachability (e.g.~\cite{bansal2017hamilton, ames2019control, gurriet2019scalable, bajcsy2019efficient, choi2021robust, tonkens2022refining}) or learning-based~(e.g.~\cite{dawson2022safe, dawson2022learning, so2023train, liu2023safe, lavanakul2024safety}) can be used. However, when the environment is sensed online (and therefore the safe set is constructed online), the assumptions of a \ac{CBF} might be difficult to verify. If unverified, these controllers could fail to maintain safety. 

The third common approach is to encode safety constraints jointly between the controller and the path planner. For example, \ac{MPC} plans trajectories considering the dynamics of the robotic system, and also determines a control input to track the trajectory. Various versions of this basic concept exist, e.g.~\cite{rawlings2017model, rosolia2017autonomous, tordesillas2021faster, zhou2021raptor}. However, given the nonlinearity of the robot dynamics and the nonconvexity of the environment, guaranteeing convergence, stability or recursive feasibility is challenging. To handle the interaction between path planners and controllers,  multirate controllers~\cite{rosolia2017autonomous, agrawal2021constructive} have also been proposed. These methods exploit the differential flatness of the system to provide theoretical guarantees, although the resulting mixed-integer problem can be expensive for clutter/complicated environments. In general, these methods solve the path planning problem and the control problem separately, but impose additional constraints on each to guarantee that the robot will remain safe. This assumes a structure in the path planner and the controller, limiting the applicability. 

There is also a growing literature on end-to-end learning based methods for safe perception, planning, and control. See for e.g., \cite{chou2022safe, karkus2019differentiable} and references within. These methods can perform well in scenarios that they have been trained on, but do not provide guarantees of performance or safety in scenarios beyond which they have been trained.

The idea of backup planners/controllers has been introduced recently to address some of the above challenges. In~\cite{tordesillas2021faster}, a backup trajectory is constructed using a linear model to ensure the trajectory lies within the known safe set at all times. However, since the backup trajectory was generated using simplified dynamics, the nonlinear system may not be able to execute this trajectory, possibly causing safety violations.  A similar approach is proposed in~\cite{kousik2020bridging} for mobile robots with the ability to stop. In~\cite{singletary2022onboard}, safety is guaranteed by blending the nominal and backup control inputs. The mixing fraction is determined by numerically forward propagating the backup controller. However, due to the mixing, the nominal trajectory is never followed exactly, even when it is safe to do so. By combining elements from these methods in a novel manner, \gatekeeper{} addresses the respective limitations, without requiring the path planner and controller to be co-designed.

\section{MOTIVATING EXAMPLE and METHOD OVERVIEW}
\label{sec:motivation}

\begin{figure}[!t]
    \centering
    \includegraphics[width=\linewidth]{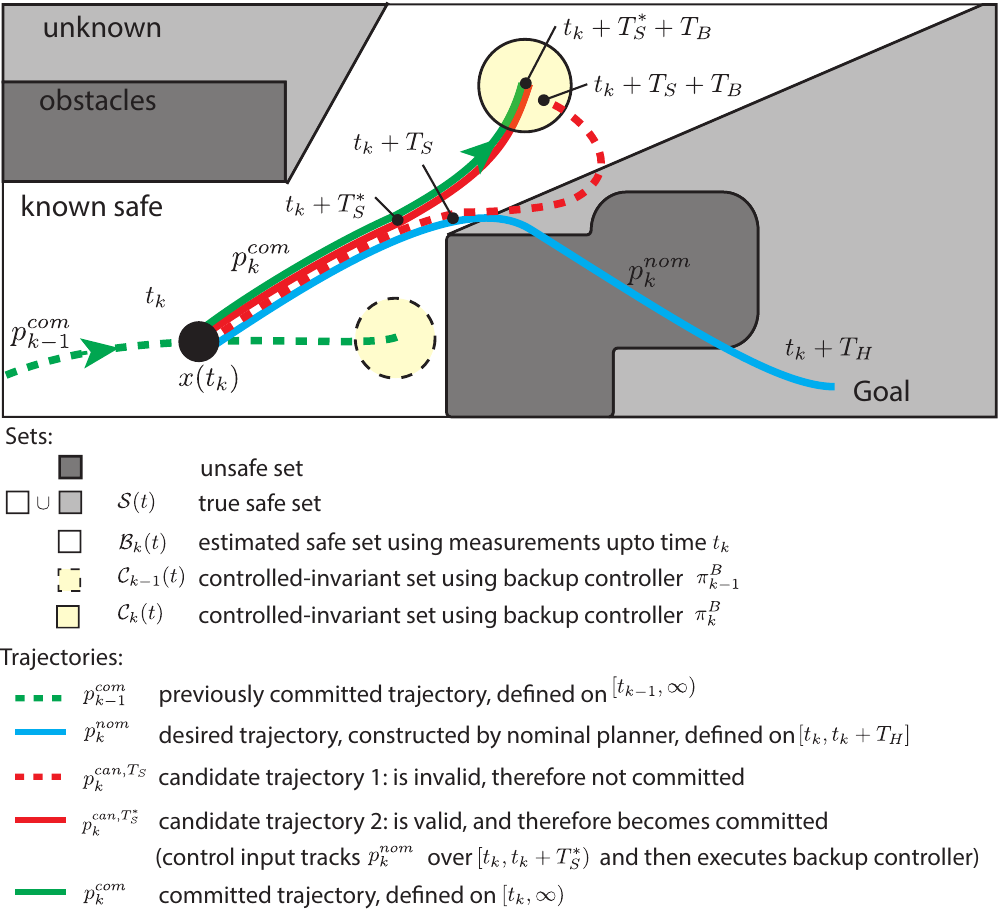}
    \caption{Notation used in this paper. The nominal planner can plan trajectories into unknown spaces, but \gatekeeper{} ensures the committed trajectory lies within the estimated safe sets, for all future time.}
    \label{fig:times}
\end{figure}

We present an example to illustrate the key concepts in this paper, and challenges when dealing with dynamic environments and limited sensing. A common wildfire fire-fighting mission is the ``firewatch" mission, where a helicopter is deployed to trace the fire-front, the outer perimeter of the wildfire. The recorded GPS trace is then used to create a map of the wildfire, which is then used to efficiently deploy appropriate resources. Today, the helicopters used in the firewatch mission are human-piloted, but in this example, we design an autonomous controller for a UAV to trace the fire-front without entering or being surrounded by the fire. Fig.~\ref{fig:times} depicts the notation used in this paper. 

The fire is constantly evolving, and expanding outwards. Thus the safe set, the set of states located outside the fire, is a time-varying set denoted $\Scal(t)$. Since the rate of spread of fire is different at each location, (it depends on various environmental factors like slope, vegetation and wind~\cite{rothermel1972mathematical, andrews2018rothermel}), the evolution of the safe set $\Scal(t)$ is unknown. That said, it is often possible to bound the evolution of $\Scal(t)$. In this example, we assume the maximum fire spread rate is known. To operate in this dynamic environment, the UAV makes measurements, for example thermal images that detect the fire-front. However, due to a limited field-of-view, only a part of the safe set can be measured.

The challenge, therefore, is to design a controller for the nonlinear system that uses the on-the-fly measurements to meet mission objectives, while ensuring the system state $x(t)$ remains within the safe set at all times, i.e., 
\eqn{
x(t) \in \Scal(t), \ \forall t \geq t_0. \label{eqn:safety}
}

Since $\Scal$ is unknown, verifying~\eqref{eqn:safety} directly is not possible. We ask a related question: given the information available upto some time $t_k$, does a candidate trajectory $p_k^{can}(t)$ satisfy
\eqn{
p_k^{can}(t) \in \Bcal_k(t), \ \forall t \geq t_k, \label{eqn:safety_bcal}
}
where $\Bcal_k(t)$ is the \emph{perceived} safe set for any time $t \geq t_k$ constructed using the sensory information available up to $t_k$ only. If we assume the perception system provides a reliable estimate of a subset of the safe set, $\Bcal_k(t) \subset \Scal(t) \ \forall t \geq t_k$, then any candidate trajectory satisfying~\eqref{eqn:safety_bcal} will also satisfy $p_k^{can}(t) \in \Scal(t)$. However, since the check in~\eqref{eqn:safety_bcal} needs to be performed over an infinite horizon $t \geq t_k$, it still cannot be implemented. \emph{A key contribution of this paper is to show how we can perform this check by verifying only a finite horizon.}

We propose the following: at each iteration, we construct a \emph{candidate trajectory} and check whether the candidate satisfies~\eqref{eqn:safety_bcal}. If so, the candidate trajectory becomes a \emph{committed trajectory}. The controller always tracks the last committed trajectory, thus ensuring safety. In other words, the \emph{candidate trajectory} is \emph{valid} if it is safe over a finite horizon and reaches a backup set by the end of the horizon. The controller tracks the last valid trajectory (i.e., the committed trajectory), until a new valid trajectory is found. 

Referring back to the firewatch mission, if the UAV is able to fly faster than the maximum spread rate of the fire, a safe course of action could be to simply fly perpendicular to the firefront, i.e., radially from the fire faster than the maximum fire spread rate. This is an example of a \emph{backup controller}, since it encodes the idea that if the system state reaches a backup set $\Ccal_k(t_{kB})$ at some time $t_{kB} \geq t_k$, then the backup controller $\pi_B^k$ will ensure that $x(t) \in \Ccal_k(t)$ for all $t \geq t_{kB}$. Note, the notation $\Ccal_k(t)$ highlights that the backup set could be a time-varying set. This switching time $t = t_k + T_s$ will be maximized by \gatekeeper{} since it is off-nominal behavior. 

In the firewatch mission, $\pi_B^k$ is controller to make the UAV fly perpendicular to the firefront, and $\Ccal_k(t)$ is the set of states that are ``sufficiently far from fire, with a sufficiently high speed perpendicular to the fire." A worked example with exact expressions for $\Scal(t), \Bcal_k(t), \Ccal_k(t)$  is provided in the appendix. Since the fire is constantly expanding, the $\Ccal_k(t)$ set is also time-varying: the set of safe states needs to be moving radially outwards. Furthermore, at each $k$, the backup controller and set can be a different, so we index these by $k$ too.

Using  backup controllers, we can find a sufficient condition for~\eqref{eqn:safety_bcal} that only requires finite horizon trajectories:
\eqn{
&\begin{cases}
    p_k^{can}(t) \in \Bcal_k(t) & \text{ if } t \in [t_k, t_{kB})\\
    p_k^{can}(t_{kB}) \in \Ccal_k(t_{kB})
\end{cases}\label{eqn:main_safety_check}\\
 &\implies \begin{cases}
    p_k^{can}(t) \in \Scal(t) & \text{ if } t \in [t_k, t_{kB})\\
    p_k^{can}(t) \in \Scal(t) & \text{ if } t \in [t_{kB}, \infty)
\end{cases}\\
&\iff p_k^{can}(t) \in \Scal(t) \ \forall t \geq t_k
}
for any $t_{kB} \geq t_k$, provided (I) $\Bcal_k(t) \subset \Scal(t)$, (II)~$\Ccal_k(t) \subset \Scal(t) \ \forall t \geq t_{kB}$, and (III)~for $t \geq t_{kB}$ the control input to the  candidate trajectory is $\pi_B^k$. These conditions can be verified easily: (I)~\emph{is the assumption that the perception system correctly identifies a subset of the safe set}, (II)~is the defining property of a backup set, and (III)~will be true based on how we construct the candidate trajectory. 

Notice that in~\eqref{eqn:main_safety_check}, we only need to verify the candidate trajectory over a finite interval $[t_k, t_{kB}]$, but this is sufficient to proving that the candidate is safe for all $t \geq t_{k}$. 

In the following sections, we formalize the \gatekeeper{} as a method to construct safe trajectories that balance between satisfying mission objectives and ensuring safety.

\section{PROBLEM FORMULATION}
\label{sec:problem}

\begin{table}[t]
   \centering
   \topcaption{Notation}
   \begin{tabular}{@{} c p{2.7in} @{}}
      \toprule
      Symbol & Definition\\
      \midrule
      \multicolumn{2}{l}{Time Points:} \\ 
      $t_k$ & Start time of iteration $k$\\
      $t_{kS}$ & Switch time $t_{kS} = t_k + T_S$\\
      $t_{kB}$ & Forecast time $t_{kB} = t_{kS} + T_B$\\
      \midrule
      \multicolumn{2}{l}{Sets:} \\ 
      $\Xcal \subset \R^n$ & State space\\
      $\Ucal \subset \R^m$ & Control input space \\
      $\Scal(t) \subset \Xcal$ & Safe set at time $t$ \\
      $\Bcal_k(t) \subset \Xcal$ & Perceived safe set at time $t$ based on measurements upto time $t_k \leq t$\\
      $\Ccal_k(t) \subset \Xcal$ & $k$-th controlled-invariant set\\
      \midrule
      \multicolumn{2}{l}{Controllers:}\\
      $\pi_T$ & Trajectory tracking controller, $\pi_T: \Xcal \times \Xcal \to \Ucal$\\
      $\pi_B$ & Backup controller, $\pi_B: \R \times \Xcal \to \Ucal$\\
      \midrule
      \multicolumn{2}{l}{Trajectories:} \\ 
      $p_k^{nom}$ & $k$-th nominal trajectory\\
      $p_k^{can}$ & $k$-th candidate trajectory\\
      $p_k^{com}$ & $k$-th committed trajectory\\

      \bottomrule
   \end{tabular}
   \label{tab:nomenclature}
\end{table}

We consider two types of systems: (A)~a nominal system, with perfect state information and without disturbances, and (B)~a perturbed estimate-feedback system, where there are bounded disturbances on both the system dynamics and the measurements, and an observer estimates the state. 

\subsection{Nominal System Description}

Consider a nonlinear system,
\eqn{
\label{eqn:dynamics}
\dot x &= f(x, u)
}
where $x \in \Xcal \subset \R^n$ is the state and $u \in \Ucal \subset \R^m$ is the control input. $f: \Xcal \times \Ucal \to \R^n$ is assumed locally Lipschitz. 

Given a control policy $\pi: [t_0, \infty) \times \Xcal \to \Ucal$ and an initial condition $x(t_0) = x_0 \in \Xcal$, the initial-value problem describing the (nominal) closed-loop system is:
\eqn{
\dot x &= f(x, \pi(t, x)),  &x(t_0) = x_0. \label{eqn:closed_loop}
}
When $\pi$ is piecewise continuous in $t$ and Lipschitz wrt $x$, there exists an interval over which the solutions of~\eqref{eqn:closed_loop} exist and are unique~\cite[Thm 3.1]{khalil2002nonlinear}. We assume this interval is $[t_0, \infty)$. 

\subsection{Perturbed System Description}
Now consider a perturbed system without perfect state information. The perturbed system dynamics are
\begin{subequations}
\label{eqn:dynamics_perturbed}
\eqn{
  \dot x &= f(x, u) +  d(t),\\
  y & = c(x) + v(t), \label{eqn:measurement}
}
\end{subequations}
where $y \in \R^p$ is the sensory output, and $c: \Xcal \to \R^p$ is locally Lipschitz continuous. The additive disturbances $d: [t_0, \infty) \to \R^n$ and $v: [t_0, \infty) \to \R^p$ are bounded, $\sup_{t\geq t_0} \norm{d(t)} = \bar d < \infty$, $\sup_{t\geq t_0} \norm{v(t)} = \bar v < \infty$.

An observer-controller uses a state estimate $\hat x \in \Xcal$ to compute the control input, and takes the form
\begin{subequations}
\label{eqn:observer-controller}
\eqn{
\dot{\hat x} &= q(\hat x, y, u)\\
u &= \pi(t, \hat x)
}
\end{subequations}
where $q : \Xcal \times \R^p \times \Ucal \to \R^n$ is locally Lipschitz in all arguments. The estimate-feedback controller $\pi: \Rnonneg \times \Xcal \to \Ucal$ is assumed piecewise-continuous in $t$ and Lipschitz in $\hat x$.

In this case, the closed-loop system dynamics are:
\begin{subequations}
\label{eqn:closed_loop_perturbed}
\eqn{
\dot {\hat x} &= q(\hat x, y, \pi(t, \hat x)), &\hat x(t_0) = \hat x_0,\\
\dot x &= f(x, \pi(t, \hat x)) + d(t),  &x(t_0) = x_0\\
y &= c(x) + v(t)
}
\end{subequations}
We assume that for each initial $(x_0, \hat x_0)$ and disturbance signals $d, v$, a unique solution exists for all $t \in [t_0, \infty)$.

\subsection{Set Invariance}
Our method is based on concepts in set invariance.
\begin{definition}[Controlled-Invariant Set]
    For the nominal system~\eqref{eqn:dynamics}, a controller $\pi: [t_0, \infty) \times \Xcal \to \Ucal$ \emph{renders a set $\Ccal(t) \subset \Xcal$ controlled-invariant} on $t_0$ if, for the closed-loop system~\eqref{eqn:closed_loop} and any $\tau \geq t_0$,
    \eqn{
    x(\tau) \in \Ccal(\tau) \implies x(t) \in \Ccal(t), \ \forall t \geq \tau.
    }
\end{definition}
The concept of controlled invariance can be extended to the case with disturbances and an observer-controller~\cite{agrawal2022safe}.
\begin{definition}[Robustly Controlled-Invariant Set]
    For the perturbed system~\eqref{eqn:dynamics_perturbed}, an observer-controller~\eqref{eqn:observer-controller} \emph{renders a set $\Ccal(t) \subset \Xcal$ robustly controlled-invariant} on $t_0$ if, for the closed-loop system~\eqref{eqn:closed_loop_perturbed} and any bounded disturbance $d, v$ with $\sup_{t \geq t_0} \norm{d(t)} \leq \bar d$,  $\sup_{t \geq t_0} \norm{v(t)} \leq \bar v$, for any $\tau \geq t_0$,
    \eqn{
    x(\tau) \in \Ccal(\tau), \norm{\hat x(\tau) - x(\tau)} \leq \delta \implies x(t) \in \Ccal(t), \ \forall t \geq \tau.
    }
    for some $\delta > 0$. 
\end{definition}

Usually, the objective is to the find the largest controlled-invariant set $\Ccal(t)$ for a given safe set $\Scal(t)$, referred to as the viability kernel~\cite{blanchini1999set, gurriet2018towards, choi2021robust}. However, these methods are difficult to apply when the safe set $\Scal(t)$ is unknown apriori, and instead is estimated online. The objective and approach of this paper is different, as described below. 

\subsection{Assumptions}
\label{sec:assumptions}

Here, we formally state the assumptions that will be used to prove that \gatekeeper{} renders a system safe. We assume the following modules are available, and explain the technical assumptions of each in the following paragraphs.
\begin{enumerate}
    \item a perception system that can sense the environment, and can estimate the safe set, 
    \item a nominal planner that generates desired trajectories to satisfy mission requirements (for example reaching a goal state, or exploring a region), potentially using simplified dynamic models,
    \item an input-to-state stable tracking observer-controller that can robustly track a specified trajectory,
    \item a backup control policy that can stabilize the system to a control invariant set. 
\end{enumerate}

More specifically:

\subsubsection{Perception System}
The (potentially time-varying) safe set is denoted $\Scal(t) \subset \Xcal$. We assume $\Scal(t)$ always has a non-empty interior. Although the full safe set may not be known at any given time, using sensors and a model of the environment, there are scenarios in which it is possible to construct reasonable bounds on the evolution of the safe set. For example, in the firefighting scenario, an upper-bound on the fire's spread rate could be known. Similarly, in an environment with dynamic obstacles, we assume that a reasonable upper-bound on the velocity or acceleration of the dynamic obstacles is known. As such, although we address safety in unknown environment, we still require some assumptions on the behavior of the environment to guarantee safety. 

Specifically, we assume that the perception system provides \emph{estimates} of the safe set that are updated as new information is acquired by the sensors. The information is available at discrete times $t_k$, $k \in \naturals$. Let $\Bcal_k(t)$ denote the perceived safe set for time $t \geq t_k$ constructed using sensory information upto time $t_k$. We assume the following:
\begin{assumption}
\label{assump:bcal}
The safe set $\Scal(t) \subset \Xcal$ has a non-empty interior for each $t$, and the estimated safe set $\Bcal_k(t)$ satisfies
\begin{subequations}
\label{eqn:bcal_assumptions}
\eqn{
&\Bcal_{k}(t) \subset \mathcal S(t) &&\forall k \in \naturals, t \geq t_k, \label{eqn:bcal_ass_1}\\
&\Bcal_{k}(t) \subset \Bcal_{k+1}(t) && \forall k\in\naturals, t \geq t_{k+1}. \label{eqn:bcal_ass_2}
}
\end{subequations}
\end{assumption}
This reads as follows. In~\eqref{eqn:bcal_ass_1}, we assume that any state perceived to be safe is indeed safe. In \eqref{eqn:bcal_ass_2}, we assume that the predictions are conservative, i.e., new information acquired at $t_{k+1}$ \emph{does not} reclassify a state $x \in \Bcal_k(t)$ (i.e. a state perceived to be safe based on information time $t_k$) as an unsafe state $x \notin \Bcal_{k+1}(t)$ based on information received at $t_{k+1}$.

This assumption (while stated more generally) is common in the literature on path planning in dynamic/unknown environments~\cite{tordesillas2021mader, zhou2021raptor}. Depending on the application, various methods can be used to computationally represent such sets, including SDFs~\cite{oleynikova2017voxblox} or SFCs~\cite{liu2017planning}.  If there are perception or predictions uncertainties, we assume they have already been accounted for when constructing $\Bcal_k(t)$. Some methods to handle such errors are studied in~\cite{agrawal2024online} and references therein.

Note, Assumption~\ref{assump:bcal} \emph{does not} require that if a state $x$ is classified as safe at some time $t_k$, that $x$ is safe for all time. Mathematically, we \emph{do not} assume $x \in \Bcal_k(t) \implies x \in \Bcal_k(\tau) \ \forall \tau \geq t$. In the appendix, diagrams and a worked example with the firefighting mission is provided to help clarify Assumption~\ref{assump:bcal} and the definitions of $\Scal(t), \Bcal_k(t)$. 

\subsubsection{Nominal Planner}
We assume that a nominal planner enforces the mission requirements by specifying the desired state of the robot for a short horizon $T_H$ into the future. 

\begin{definition}[Trajectory]
    A \emph{trajectory} $p$ with horizon $T_H$ is a piecewise continuous function $p : \Tcal \to \Xcal$ defined on $\Tcal = [t_k, t_k + T_H] \subset \R$. 
    A trajectory $p$ is \emph{dynamically feasible} wrt~\eqref{eqn:dynamics} if there exists a piecewise continuous control  $u : \Tcal \to \Ucal$ s.t. 
    \eqn{
    p(t) = p(t_k) + \int_{t_k}^{t} f(p(\tau), u(\tau)) d\tau, \quad    \forall t \in \Tcal.
    }
\end{definition}

Denote the nominal trajectory available at the $k$-th iteration by the function $p^{nom}_k: [t_k, t_k + T_H] \to \Xcal$. We do not require $p^{nom}_k$ to be dynamically feasible wrt~\eqref{eqn:dynamics} or~\eqref{eqn:dynamics_perturbed}.

Note, although some path planners (e.g. A*, RRT*) construct geometric paths, we assume the output of the path planner is a trajectory, i.e., is parameterized by time. Methods for time allocation of geometric paths is a well studied problem, see for e.g.~\cite{richter2016polynomial, liu2017planning, tordesillas2021faster}. 

To summarize, we assume a nominal planner is available:
\begin{assumption}
\label{assumption:nom_planner}
There exists a nominal planner that can generate finite-horizon trajectories $p_k^{nom}: [t_k, t_k+T_H] \to \Xcal$ for each $k \in \naturals$. 
\end{assumption}

\subsubsection{Tracking Observer-Controller}
We assume an estimate-feedback controller $\pi_T: \Xcal \times \Xcal \to \Ucal$ that computes a control input $u = \pi_T(\hat x, p(t))$ to track a given trajectory $p$; we refer to this policy as the tracking observer-controller~\cite{kanayama1990stable, lee2010geometric, martin2003flat}. We assume that the tracking controller is input-to-state stable~\cite{agrawal2022safe}:

\begin{definition}[Input-to-State Stable Observer-Controller]
\label{def:pi_t}
Let $\Tcal = [t_k, t_l] \subset \Rnonneg$. A tracking observer-controller
\begin{subequations}
\eqn{
u(t) &= \pi_T(\hat x, p(t))\\
\dot {\hat x} &= q(\hat x, y, u)
}
\end{subequations}
is \emph{input-to-state stable} for the system~\eqref{eqn:dynamics}, if, for any bounded disturbances $d: \Tcal \to \R^n$, $v: \Tcal \to \R$, and any dynamically feasible trajectory $p: \Tcal \to \Xcal$, the following holds:
\eqn{
&\norm{x(t_k) - \hat x(t_k)} \leq \delta, \text{ and } \quad p(t_k) = \hat x(t_k) \implies \notag \\
&\quad  \norm{x(t) - \hat x(t)} \leq \beta(\delta, t-t_k) + \gamma(\bar w), \text{ and } \notag \\ 
&\quad \norm{\hat x(t) - p(t)} \leq \beta(\delta, t-t_k) + \gamma(\bar w) , \text{ and } \notag \\
&\quad \norm{x(t) - p(t)} \leq \beta(\delta, t-t_k) + \gamma(\bar w), \ \forall t \in \Tcal, \label{eqn:disturbance_state_stable}
}
where $\beta: \Rnonneg \times \Rnonneg \to \Rnonneg$ is \classKL, $\gamma : \Rnonneg \to \Rnonneg$ is \classK, and $\bar w = \max(\sup_{t \in \Tcal}{\norm{d(t)}}, \sup_{t \in \Tcal}{\norm{v(t)}})$.
\end{definition}
Note, for simplicity we assumed the same $\beta, \gamma$ for each of the three norms in~\eqref{eqn:disturbance_state_stable}, although it is not strictly necessary. 

To summarize, we assume a tracking controller is known:
\begin{assumption}
\label{assum:pi_t}
There exists an input-to-state stable observer-controller of the form in Def.~\ref{def:pi_t}, with known functions $\beta, \gamma$.
\end{assumption}

\subsubsection{Backup Controller}
In the case when a safe set $\mathcal S$ can not be rendered controlled invariant for given system dynamics, the objective reduces to finding a set $\Ccal \subset \Scal$, and a controller $\pi: \Ccal \to \Ucal$ that renders $\Ccal$ controlled invariant. 
For example, by linearizing~\eqref{eqn:dynamics} around a stabilizable equilibrium $x_{e}$, an LQR controller renders a (sufficiently small) set of states around $x_{e}$ forward invariant~\cite[Thm. 4.13, 4.18]{khalil2002nonlinear}. This observation leads to the notion of backup safety~\cite{chen2021backup, singletary2022onboard}.

\begin{definition}[Backup Controller]
    A controller $\pi_B^k: \Tcal \times \Xcal \to \Ucal$ is a \emph{backup controller} to a set $\Ccal_k(t) \subset \Xcal$ defined for $t \in \Tcal = [t_k, \infty)$ if, for the closed-loop system
    \eqn{
    \dot x = f(x, \pi_B^k(t, x)),
    }
    (A) there exists a neighborhood $\Ncal_k(t) \subset \Xcal$ of $\Ccal_k(t)$, s.t. $\Ccal_k(t)$ is reachable in fixed time $T_B$:
    \eqn{
    x(\tau) \in \Ncal_k(\tau) \implies x(\tau + T_B) \in \Ccal(\tau + T_B),
    }
    and (B) $\pi_B^k$ renders $\Ccal_k(t)$ controlled-invariant:
    \eqn{
    x(\tau + T_B) \in \Ccal(\tau + T_B) \implies x(t) \in \Ccal(t) \ \forall t \geq \tau + T_B.
    }
\end{definition}

\begin{remark}
The neighborhood $\Ncal_k$ does not need to be known in the \gatekeeper{} framework. The definition ensures that there exist states outside $\Ccal_k$ that can be driven into $\Ccal_k$ by the backup controller within a fixed time $T_B$. This excludes cases, for example, where the backup trajectories approach $\Ccal_k$ asymptotically but never actually enter $\Ccal_k$. 
\end{remark}

To summarize, we assume a backup controller is known:
\begin{assumption}
\label{assum:ccalk}
    At the $k$-th iteration, a set $\Ccal_k(t)$ and a backup controller $\pi_B^k: [t_k, \infty) \times \Xcal \to \Ucal$ to $\Ccal_k(t)$ can be found where
    \eqn{
    \Ccal_k(t) \subset \Scal(t), \ \forall t \geq t_k. \label{eqn:ccal_assmp}
    }
\end{assumption}

\remark{Note that while we assume $\Ccal_k(t) \subset \Scal(t)$, we \emph{ do not } assume the trajectory to reach $\Ccal_k(t)$ is safe, nor that the set $\Ccal_k(t)$ is reachable from the current state $x(t_k)$ within a finite horizon. This is in contrast to backward reachability based methods~\cite{abate2020enforcing, llanes2021safety, tonkens2022refining, gurriet2019realizable}. Instead, we will ensure both of these conditions are satisfied through our algorithm.}

\begin{remark}
The design of backup controllers and sets depends on the robotic system and the environment model. For some systems, the backup set can be designed by linearization about a stabilizable equilibrium point (or limit cycle), and determining the region of attraction. Other methods include reachability analysis or learning-based approaches, e.g.~\cite{bansal2017hamilton,dawson2022learning, dawson2022safe, so2023train, liu2023safe, lavanakul2024safety}. Generic methods to design the backup controllers are beyond the scope of this paper, but specific methods are discussed in Section~\ref{sec:sims_and_exps} and in~\cite{chen2021backup, singletary2022onboard}.
\end{remark}

\subsection{Problem Statement}
In summary, the problem statement is
\begin{problem}
  Consider system~\eqref{eqn:dynamics_perturbed} satisfying Assumptions~\ref{assump:bcal}-\ref{assum:ccalk}, i.e., a system with a perception system satisfying Assumption~\ref{assump:bcal}, a nominal planner that generates desired trajectories, an input-to-state stable tracking controller satisfying Definition~\ref{def:pi_t}, and a backup controller satisfying Assumption~\ref{assum:ccalk}. Design an algorithm to track desired trajectories while ensuring safety, i.e., $x(t) \in \Scal(t)$ for all $t \geq t_0$. 
\end{problem}

\section{PROPOSED SOLUTION}
\label{sec:solution}

\gatekeeper{} is a module that lies \emph{between} the planning and control modules. It considers the nominal trajectories by the planner, modifies them as needed to what we call committed trajectories, and inputs these committed trajectories to the trajectory-tracking controller. In this section, we will demonstrate how to construct these committed trajectories. To aid the reader, the analysis is first presented for the nominal case, and later extended to the perturbed case. The various trajectories and times are depicted in Fig.~\ref{fig:times}. The algorithm is described in Algorithm~\ref{alg:full} and depicted in Fig.~\ref{fig:block_diagram}.

\subsection{Nominal Case}
At the $k$-th iteration, $k\in\naturals\setminus\{0\}$, let the previously committed trajectory be $p^{com}_{k-1}$. \gatekeeper{} constructs a candidate trajectory $p_k^{can, T_S}$ by forward propagating a controller that tracks $p^{nom}_k$ over an interval $[t_k, t_k + T_S)$, and executes the backup controller for $t \geq t_k + T_S$. $T_S \in \Rnonneg$ is a switching duration maximized by \gatekeeper{} as described later. Formally, 
\begin{definition}[Candidate Trajectory]
    Suppose at $t=t_k$, 
    \begin{itemize}
        \item the state is $x(t_k) = x_k$,
        \item the nominal trajectory is $p^{nom}_k: [t_k, t_k+T_H] \to \Xcal$,
        \item $\pi_T$ is a trajectory tracking controller,
        \item $\pi_B^k$ is a backup controller to the set $\Ccal_k(t)$.
    \end{itemize}
    Given a $T_S \in [0, T_H]$, the \emph{candidate trajectory} $p^{can, T_S}_{k}: [t_k, \infty) \to \Xcal$ is the solution to the initial value problem 
    \begin{subequations}
    \label{eqn:pcan}
    \eqn{
    \dot p &= f(p, u(t)),\\
    p(t_k) &= x_k, \\
    u(t) &= \begin{cases}
        \pi_T(p(t), p^{nom}_k(t)) & t \in [t_k, t_k + T_S)\\
        \pi_B^k(t, p(t)) & t \geq t_k + T_S.
    \end{cases}
    }
    \end{subequations}
\end{definition}

By construction, the candidate is dynamically feasible wrt~\eqref{eqn:dynamics}. A candidate trajectory is \emph{valid} if the following hold:
\begin{definition}[Valid]
\label{def:valid}
    A candidate trajectory $p^{can, T_s}_k: [t_k, \infty) \to \Xcal$ defined by~\eqref{eqn:pcan} is \emph{valid} if 
 the trajectory is safe wrt the estimated safe set over a finite interval:
        \eqn{
        p^{can, T_S}_{k}(t) \in \Bcal_k(t), \ \forall t \in [t_k, t_{k, B}],
        \label{eqn:validate_path}
        }
  and the trajectory reaches $\Ccal_k(t)$ at the end of the horizon:
        \eqn{
        p^{can, T_S}_{k}(t_{k, B}) \in \Ccal_k(t_{k, B}), \label{eqn:validate_end}
        }
    where $t_{k, B} = t_k + T_S + T_B$.
\end{definition}

Notice checking whether a candidate is valid only requires the solution $p_k^{can, T_S}$ over the finite interval $[t_k, t_k + T_S + T_B]$. This means that the candidate can be constructed by numerical forward integration over a finite horizon. 

Def.~\ref{def:pcom} defines how the $k$-th committed trajectory is constructed using the nominal trajectory $p_k^{nom}$, the backup controller $\pi_k^B$, and the previous committed trajectory $p_{k-1}^{com}$. 
\begin{definition}[Committed Trajectory]
\label{def:pcom}
At the $k$-th iteration, define 
\eqn{
\Ical_k = &\Big\{ T_S \in [0, T_H]: p^{can, T_S}_{k} \text{ is valid} \Big\} \subset \R,
}
where $p^{can, T_S}_{k}: [t_k, \infty) \to \Xcal$ is as defined in~\eqref{eqn:pcan}, and Def.~\ref{def:valid} is used to check validity. The committed trajectory is $p^{com}_k: [t_k, \infty) \to \Xcal$, defined as follows:

If $\Ical_k \neq \emptyset$, let $T_S^* = \max \Ical_k$. The \emph{committed trajectory} is 
\eqn{
p^{com}_k(t) = p^{can, T_S^*}_k(t), 
\quad t\in[t_k,\infty).
}

If $\Ical_k = \emptyset$, the \emph{committed trajectory} is
\eqn{
\label{eqn:dont_update}
p^{com}_k(t) = p^{com}_{k-1}(t), \quad t\in[t_k,\infty).
}
\end{definition}

We are ready to prove the proposed strategy guarantees safety. First, we show that each committed trajectory is safe. 
\begin{theorem}
\label{theorem:safety}
    Suppose Assumptions~\ref{assump:bcal}-\ref{assum:ccalk} hold. Suppose $p^{can, T_S}_0 : [t_0, \infty) \to \Xcal$ is a candidate trajectory that is dynamically feasible wrt~\eqref{eqn:dynamics} and valid according to Def.~\ref{def:valid} for some $T_S \geq 0$. If, for every $k \in \naturals$, $p^{com}_{k} :[t_k, \infty) \to \Xcal$ is determined using Def.~\ref{def:pcom}, then for all $k \in \naturals$, 
    \eqn{
    p^{com}_k(t) \in \Scal(t), \quad \forall t \in [t_k, \infty).
    }
    \end{theorem}
    
    \begin{proof}
    The proof is by induction.
    
 \emph{Base Case: $k=0$. } Since $p_0^{can}$ is a valid trajectory, it is committed, i.e., $p_0^{com} = p_0^{can, T_S}$. Then, 
    \eqnN{
    &p_0^{com}(t) \in \begin{cases}
    \Bcal_0(t) & \text{for } t \in [t_0, t_{0, B})\\
    \Ccal_0(t) & \text{for } t = t_{0, B}
    \end{cases}\\ 
    &\implies p_0^{com}(t) \in \begin{cases}
    \Scal(t) & \text{for } t \in [t_0, t_{0, B})\\
    \Scal(t)& \text{for } t \geq t_{0, B}
    \end{cases}\\
    &\iff p_0^{com}(t)\in \Scal(t)\  \text{for } t \geq t_0
    }
    where $t_{0, B} = t_0 + T_S + T_B$. 

    \emph{Induction Step:} Suppose the claim is true for some $k\in\naturals$. We will show the claim is also true for $k+1$. There are two possible definitions for $p_k^{com}$: 
    
    \emph{Case 1:} When $\Ical_{k+1} \neq \emptyset$, $p_{k+1}^{can, T_S^*}$ is a valid candidate, i.e.,
    \eqnN{
    p_{k+1}^{com}(t) &= p_{k+1}^{can, T_S^*}(t) \ \forall t \geq t_0\\
    &\in \begin{cases}
    \Bcal_{k+1}(t) & \text{for } t \in [t_{k+1}, t_{k+1, SB})\\
    \Ccal_{k+1}(t) & \text{for } t \geq t_{k+1, SB}
    \end{cases}\\
    &\in \Scal(t)\  \text{for } t \geq t_{k+1}
    }
    
    \emph{Case 2:} If $\Ical_{k+1} = \emptyset$, the committed trajectory is unchanged, 
    \eqnN{
    p^{com}_{k+1}(t) = p^{com}_k(t) \in \Scal(t), \ \forall t\geq t_{k+1}.
    }
\end{proof}

The following shows that \gatekeeper{} ensures safety.
    
    \begin{theorem}    
    \label{theorem:safety_part2}
    Under the assumptions of Theorem~\ref{theorem:safety}, if $x(t_0) = p^{com}_0(t_0)$, and for each $k \in \naturals$ the control input to the nominal system~\eqref{eqn:dynamics} is
    \eqn{
    u(t) = \pi_T^k(x(t), p_k^{com}(t)), \forall t \in [t_k, t_{k+1}), \label{eqn:tracking_controller_theorem2}
    }
    then the closed-loop dynamics~\eqref{eqn:closed_loop} will satisfy 
    \eqn{
    x(t) \in \Scal(t), \forall t \geq t_0.
    }
\end{theorem}
    \begin{proof}
We prove this by showing that $\forall k\in \naturals$, $x(t) = p_k^{com}(t)$ for $t \in [t_k, t_{k+1})$. Again, we use induction.
    
    \emph{Base Case:}  For the nominal system~\eqref{eqn:dynamics}, when $x(t_0) = p_0^{com}(t_0)$ and the tracking controller is ISS~\eqref{eqn:disturbance_state_stable}, 
    \eqnN{
    \norm{x(t) - p_0^{com}(t)} \leq \beta(0, t-t_0) + \gamma(0) = 0 \\
    \therefore x(t) = p_0^{com}(t) \; \forall t \in [t_0, t_1)
    }
    
    \emph{Induction Step:} Suppose for some $k \in \naturals$, $x(t) = p_k^{com}(t)$ for $t \in [t_k, t_{k+1})$. There are two cases for $p_{k+1}^{com}$: 
    
    \emph{Case 1:} a new candidate is committed, $\therefore p_{k+1}^{can, T_S}(t_{k+1}) = x(t_{k+1})$. Since the tracking controller is input-to-state stable, this implies $x(t) = p_{k+1}^{com}(t)$ for $t \in [t_{k+1}, t_{k+2})$. 
    
    \emph{Case 2:} A new candidate is not committed, $\therefore p_{k+1}^{com}(t) = p_k^{com}(t) \ \forall t \in [t_{k+1}, t_{k+2})$. Since $x(t_{k+1}) = p_{k}^{com}(t_{k+1})$, the tracking controller ensures $x(t) = p_{k+1}^{com}(t) \ \forall t \in [t_{k+1}, t_{k+2})$. 
    
Therefore, $x(t)  = p_k^{com}(t) \in \Scal(t) \ \forall t \in [t_k, t_{k+1})$, for each $k \in \naturals$. Thus, $x(t) \in \Scal(t)$ for all $t \geq t_0$. 
\end{proof}

\begin{remark}
The controller in~\eqref{eqn:tracking_controller_theorem2} uses the backup controller $\pi_B$: the committed trajectory $p_k^{com}$ is constructed such that for all $t \geq t_{k, SB}$ the trajectory uses the backup controller (see \eqref{eqn:pcan}). Therefore, if after the $k$-th step new candidate trajectories are not committed, the controller~\eqref{eqn:tracking_controller_theorem2} applies the backup controller for time $t \geq t_{k, SB}$. 
\end{remark}

\begin{remark}
In~\cite{chen2021backup, singletary2022onboard}, numerical forward propagation of the trajectory with a backup controller is also used to construct a safety filter. However, the resulting control input mixes the nominal control input with the backup control input at all times. In contrast, in \gatekeeper{} we use a switching time to switch between implementing the nominal control input and the backup control input. This is desirable since it leads to less conservative controllers, as highlighted in section~\ref{sec:fire}. 
\end{remark}

\subsection{Perturbed Case}
\begin{figure}
    \centering
    \includegraphics[width=\linewidth]{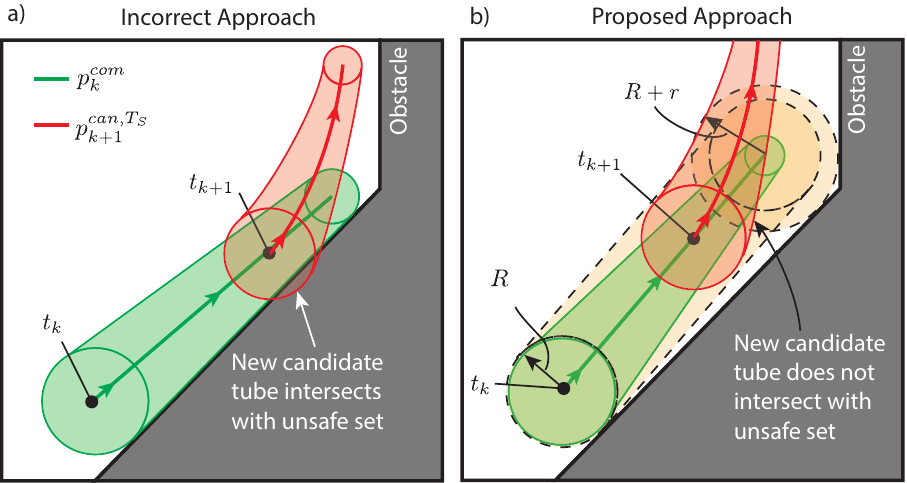}
    \caption{Diagram depicting the challenge due to disturbances. (a) Green line shows the committed trajectory at iteration $k$, and the shaded region is the tube that contains the system trajectory. If the validation step only checks that the green tube lies within the safe set, a new candidate trajectory (red) cannot be committed, since the candidate tube (red shaded region) intersects with the unsafe set. (b)~shows the proposed approach, where safety is checked wrt the yellow set, i.e., a tube of radius $R$ along the trajectory and a ball of radius $R+r$ at the end. This allows for sufficient margin to commit a new trajectory at the next iteration.}
    \label{fig:tubes}
\end{figure}

We now address the case with non-zero disturbances and state-estimation error.\footnote{ Compared to the conference version~\cite{gatekeeper_iros}, here we consider the additional uncertainty due to state estimation errors, and simplify the validation check.} The algorithm is identical to that presented above, except that the validation step will be redefined. 

First, we highlight the problem that disturbances introduce. Consider the specific scenario visualized in Fig.~\ref{fig:tubes}. To account for the disturbances, we validate safety of a tube around the candidate trajectory: using the ISS bound~\eqref{eqn:disturbance_state_stable}, a tube of decreasing radius around the committed trajectory will always contain the true state of the system. Therefore, if instead of~\eqref{eqn:validate_path}, we checked that the corresponding tube containing the candidate trajectory lies within the safe set (green tube in Fig.~\ref{fig:tubes}a), then indeed, the system will remain safe. 

However, when a new candidate is proposed at the next iteration, the new tube (red tube) intersects with the unsafe set. Thus, the new candidate cannot be committed, and an undesired deadlock is reached: $x(t) \in \Ccal_k(t)$ for all $t \geq t_{k, B}$. 

To avoid this behavior, we propose a different validity check. First, we check that a tube of radius $R$ is safe over the finite horizon, and second, we ensure $\Ccal_k$ is a (larger) distance $R+r$ away from the unsafe set, where $r, R$ are defined below. In Fig.~\ref{fig:tubes}a, this is depicted  by the yellow sets. Note, the additional $+r$ term is used to avoid the described deadlock behaviour, and is not needed to guarantee safety.

Recall Def.~\ref{def:pi_t} defines the controller's tracking error bounds. The validity check in Def.~\ref{def:valid} is replaced by the following:

\begin{definition}[Robustly Valid]
\label{def:robustly_valid}
Consider the dynamical system~\eqref{eqn:dynamics_perturbed}, with bounded disturbances $\sup_{t \geq t_k} \norm{d(t)} \leq \bar d$, and $\sup_{t \geq t_k} \norm{v(t)} \leq \bar v$. Let $\bar w = \max(\bar d, \bar v)$.  Suppose $\norm{x(t_k) - \hat x(t_k)} \leq r$ for some $k \in \naturals$. Let $R = \beta(r, 0) + \gamma(\bar w)$. 

A candidate trajectory $p^{can, T_s}_k: [t_k, \infty) \to \Xcal$ defined by~\eqref{eqn:pcan} is \emph{robustly valid} if 
    \begin{itemize}
        \item the candidate trajectory coincides with the state estimate at the initial time:
        \eqn{
        \hat x(t_k) = p_k^{can, T_s}(t_k), \label{eqn:matching_initial_condition}
        }
        \item the candidate trajectory is robustly safe over a finite interval:
        \eqn{
        p^{can, T_S}_{k}(t) \in \Bcal_k(t) \ominus \ball(R) \ \forall t \in [t_k, t_{k,B}],
        \label{eqn:robust_validate_path}
        }
        \item at the end of the interval, it reaches $\Ccal_k(t)$:
        \eqn{
        p^{can, T_S}_{k}(t_{k, SB}) \in \Ccal_k(t_{k, B}), \label{eqn:robust_validate_end}
        }
        \item and the set $\Ccal_k(t)$ is $(R+r)$ away from the unsafe set:
        \eqn{
        \Ccal_k(t) \subset \Scal(t) \ominus \ball(R + r) \ \forall t \geq t_k. \label{eqn:robust_ccal_check}
        }
    \end{itemize}
\end{definition}

If a candidate trajectory is robustly valid, it can be committed. The following theorem proves that \gatekeeper{} can render the perturbed system~\eqref{eqn:dynamics_perturbed} safe. 

\begin{theorem}
\label{theorem:robust_safety}
Suppose Assumptions~\ref{assump:bcal}-\ref{assum:ccalk} hold. Suppose $p^{com}_0: [t_0, \infty) \to \Xcal$ is a committed trajectory that is \emph{robustly} valid by Def.~\ref{def:robustly_valid} for some $r > 0, T_S \geq 0$. Suppose $\norm{x(t_0) - \hat x(t_0)} \leq r$, and $p^{com}_0(t_0) = \hat x(t_0)$. 

If, for every $k  \in \naturals \backslash \{ 0 \} $, $p^{com}_{k}: [t_k, \infty) \to \Xcal$ is determined using Def.~\ref{def:pcom} (except that validity is checked using Def.~\ref{def:robustly_valid}), and the control input to the perturbed system~\eqref{eqn:dynamics_perturbed} is
\eqnN{
u(t) = \pi_T^k(\hat x(t), p_k^{com}(t)) \quad  \forall t \in [t_k, t_{k+1}]
}
then the closed-loop system~\eqref{eqn:closed_loop_perturbed} will satisfy 
\eqn{
x(t) \in \Scal(t), \quad \forall t \geq t_0. 
}
\end{theorem}

\begin{proof}
As in Thm.~\ref{theorem:safety}, we have that for any $k \in \naturals$, 
\eqn{
p_k^{com}(t) \in \Scal(t) \quad \forall t \in [t_k, \infty).
}

We aim to prove the analog of Thm~\ref{theorem:safety_part2}, i.e., that for any $k \in \naturals$, tracking the committed trajectory $p_k^{com}(t_k)$ for $t \geq t_k$ is safe. This is proved below. 

Since $p_k^{com}$ is robustly valid, $p_k^{com}(t_k) = \hat x(t_k)$. Therefore, 
\eqnN{
\norm{x(t_k) - p_k^{com}(t_k)} = \norm{x(t_k) - \hat x(t_k)} \leq r.
}
Using~\eqref{eqn:disturbance_state_stable}, this implies that for all $t \geq t_k$,
\eqnN{
\norm{x(t) - p_k^{com}(t)} &\leq \beta(r, t-t_k) + \eta(\bar w) \\
&\leq \beta(r, 0) + \eta(\bar w)= R\\
\therefore \norm{x(t) - p_k^{com}(t)} &\leq R
}
and so by~\eqref{eqn:robust_validate_path},
\eqnN{
&p_k^{com}(t) \in \Bcal_k(t) \ominus \ball(R) && \forall t \in [t_k, t_{k, SB}] \\
\implies & \{ p_k^{com}(t) \} \oplus \ball(R)  \subset  \Bcal_k(t) && \forall t \in [t_k, t_{k, SB}]\\
\implies & x(t) \in \Bcal_k(t) && \forall t \in [t_k, t_{k, SB}]. 
}

Furthermore, since for all $t \geq t_{k} + T_S$ the committed trajectory is generated by the backup controller, and $p_k^{com}(t_{k, B}) \in \Ccal_k(t_{k,B})$, we have $p_k^{com}(t) \in \Ccal_k(t), \forall t \geq t_{k, B}$. Therefore, 
\eqnN{
x(t) \in \Ccal_k(t) \oplus \ball(R) \quad \forall t \geq t_{k, B}.
}
Putting these together, 
\eqnN{
&x(t) \in \begin{cases}
\Bcal_k(t) & \text{for } t \in [t_k, t_{k, B}]\\
\Ccal_k(t) \oplus \ball(R) & \text{for } t \geq t_{k, B}
\end{cases}\\
\implies &x(t) \in \begin{cases}
\Scal_k(t) & \text{for } t \in [t_k, t_{k, B})\\
\Scal_k(t) & \text{for } t \geq t_{k, B}
\end{cases}\\
\iff & x(t) \in \Scal(t) \quad \forall t \geq t_k.
}
\end{proof}

This proves that for any $k\in \naturals$, if $p_k^{com}$ is the committed trajectory, the system will remain safe while it is tracking $p_k^{com}$. When a new candidate trajectory that is robustly valid (by Def.~\ref{def:robustly_valid}) is found, the committed trajectory can be updated, and the system will continue to remain safe. 

\begin{remark} The theorem provides certain parameters of the nominal planner. For instance, requiring trajectories to lie in $\Bcal(t_k) \ominus \ball(R)$ corresponds to the common practice of inflating the unsafe sets by a radius $R$. The theorem shows that any $R \geq \beta(r, 0) + \gamma(\bar w)$ is sufficient.
\end{remark}

\begin{remark}
In~\eqref{eqn:robust_ccal_check}, we checked that $\Ccal_k(t)$ is at least $(R+r)$ away from the boundary of $\Scal(t)$ at all $t \geq t_k$, even though the proof of safety only requires a margin $R$. The reason we check for $(R+r)$ is to prevent the deadlock scenario discussed before: under the stated assumptions, for $t\geq t_k$,  $\norm{x(t) - \hat x(t)} \leq \beta(t-t_k) + \gamma(\bar w)$.
Therefore, if $r \geq \gamma(\bar w)$ there exists some time  $\tau = t_k + T$ where $r = \beta(T, r) + \gamma(\bar w)$ since \classKL functions are strictly decreasing wrt $t$. Thus, for $t \geq \tau$,  $p_k^{com}(t) \in \Ccal_k(t)$. Thus,
\eqnN{
x(t), \hat x(t) \in \Ccal_k(\tau) \oplus \ball(r)
}
and $\norm{x(t) - \hat x (t)} \leq r$. Thus, when validating the new candidate trajectory $p_{k'}^{can, T_s}$ they will start at least $R$ away from the boundary, i.e., there is sufficient margin for new trajectories to be committed.
\end{remark}

\begin{remark} 
In constructing candidate trajectories, we require the initial state of the candidate trajectory to coincide with the state estimate, \eqref{eqn:matching_initial_condition}. If this is not the case, an additional margin would be necessary in \eqref{eqn:robust_ccal_check} to account for this error. 
\end{remark}

\begin{remark}The construction of committed trajectories is summarized in pseudo-code in Alg.~\ref{alg:full}. $\max \Ical$ can be determined efficiently, since it is an optimization of a scalar variable over a bounded interval. We used a simple grid search with $N$ points. Therefore, upto $N$ initial value problems need to be solved, which can be done very efficiently using modern solvers~\cite{rackauckas2017differentialequations}. Using $N=10$, the median computation time was only 3.4~ms. Other strategies including log-spacing or optimization techniques could be investigated in the future.
\end{remark}

\begin{algorithm}
  \label{alg:full}
  \DontPrintSemicolon
  \caption{\gatekeeper{}}
  \SetKw{KwParams}{Parameters:}
  \SetKwProg{When}{When}{}{end}
  \SetKwProg{Fn}{function}{}{end}
  \KwParams{$N > 0 \in \naturals$}\;
  \tcp{Do a grid search backwards over the interval $[0, T_H]$:}
  \For{ $i$ in range(0, $N$):}{
    Using $\Bcal_k(t)$, identify $\Ccal_k(t)$ satisfying assum.~\ref{assum:ccalk}. \;
    $T_S = (1 - i/N)T_H$ \;
    Solve the initial value problem~\eqref{eqn:pcan} to determine $p^{can, T_S}_{k}(t)$ over the interval $[t_k, t_k + T_S + T_B]$ \;
    \If{ $p^{can, T_S}_{k}$ is robustly valid by Def.~\ref{def:robustly_valid} }{
        $p^{com}_k = p^{can, T_S}_{k}$ \;
        \Return
    }
    }
    \tcp{no candidate is valid, $\Ical = \emptyset$}
    $p^{com}_k= p^{com}_{k-1}$ \;
    \Return
\end{algorithm}

\section{SIMULATIONS AND EXPERIMENTS}
\label{sec:sims_and_exps}
Code and videos are available here: \cite{gatekeeperRepo}. We test two case studies to evaluate \gatekeeper{}, where the second case study is also performed using hardware experiments. A key strength of \gatekeeper{} is that it can be composed with existing perception, planning and control algorithms, and the various techniques used are summarized in Table~\ref{tab:methods}. The details are provided in the following paragraphs.

\begin{table*}[t]
\centering
      \small
      \caption{
Methods used in implementing \gatekeeper{} for each case study.  Details are provided in the text. }
      \begin{tabular}{@{}lll@{}}
\toprule
      & Firewatch Mission & Quadrotor Navigation\\
\midrule
Sensed Data & Image of fire & RGBD image\\
Perception Output & \ac{SDF} & \ac{SDF} + \ac{SFC}\\
Nominal Planner & \ac{MPC} & \ac{DMP} \\
Tracking Controller & PD-Controller & Geometric Controller\\
Backup Controller & Fly perpendicular to fire &  Stop and yaw\\
\bottomrule
\end{tabular}
      \label{tab:methods}
\end{table*}

\subsection{Firewatch Mission}
\label{sec:fire}

\begin{figure*}[t]
  \centering
  \includegraphics[width=\linewidth]{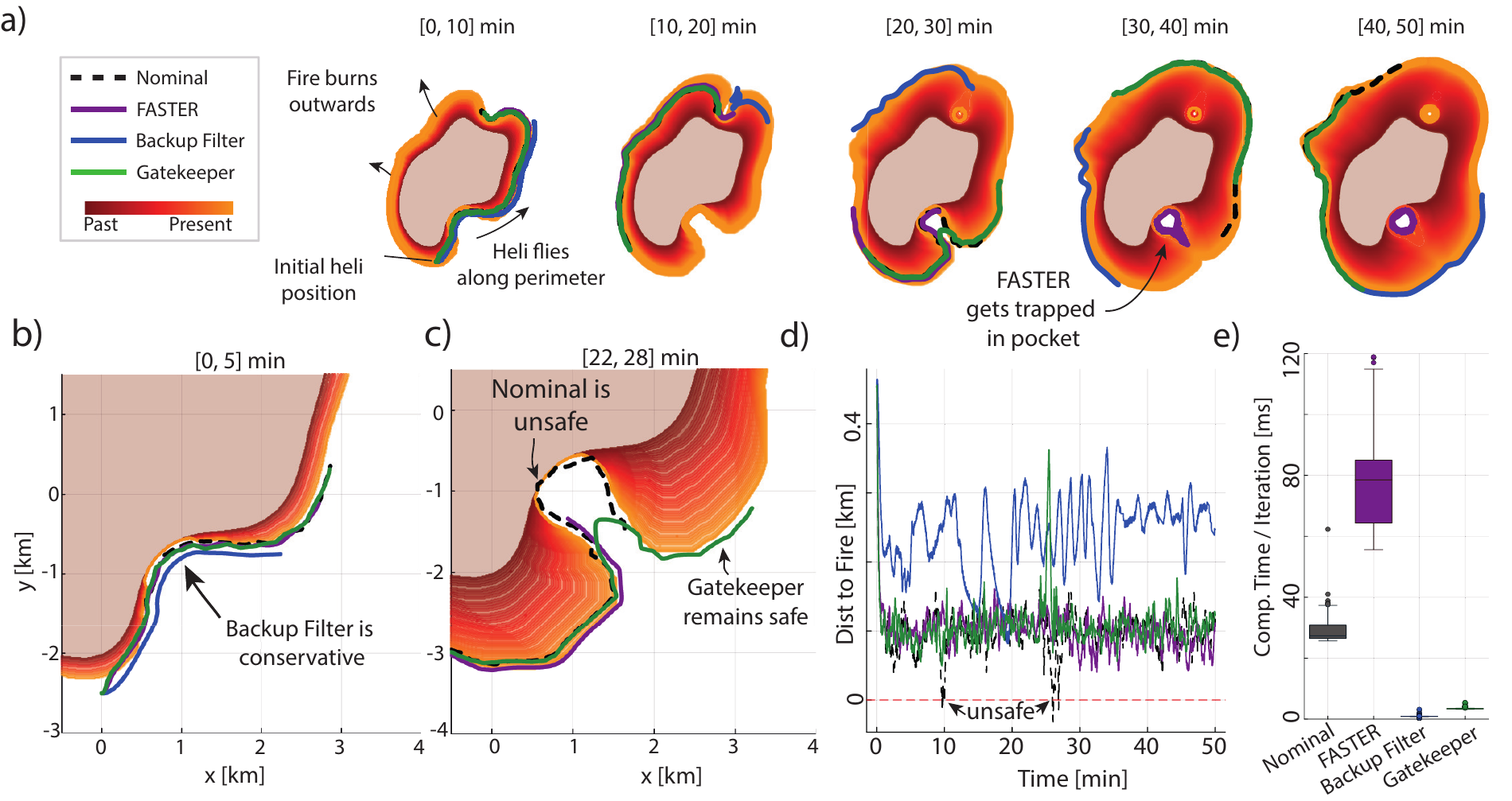}
  \label{fig:fire}
  \caption{Simulation results from Firewatch mission. (a) Snapshots of the fire and trajectories executed by each of three controller. The fire is spreading outwards, and the helicopters are following the perimeter. The black line traces the nominal controller, the blue line is based on the backup filter adapted from~\cite{singletary2022onboard} and the green line shows the proposed controller. (b, c) show specific durations in greater detail. At $t=0$, the \gatekeeper{} controller behaves identically to the nominal controller, and makes small modifications when necessary to ensure safety. The backup filter is conservative, driving the helicopter away from the fire and slowing it down. (d) Plot of minimum distance to fire-front across time for each of the controllers. (e) The nominal controller becomes unsafe 3 times, while FASTER, the backup controller, and the \gatekeeper{} controllers maintain safety. Animations are available at~\cite{gatekeeperRepo}.}
\end{figure*}

\begin{table*}[t]
      \centering
      \small
      \caption{
      Comparison of \gatekeeper{} (ours) with the nominal planner, FASTER~\cite{tordesillas2021faster}, and backup filters~\cite{singletary2022onboard}. The distance to the firefront, velocity of the helicopter, and computation time per iteration are reported for each method. IQR = interquartile range. $^*$Since the backup filter is run at each control iteration instead of every planning iteration, it runs 20~times as often as \gatekeeper{}, i.e., is 5 times as computationally expensive as \gatekeeper{}. }
      \begin{tabular}{@{}lrrrrrrrl@{}}
\toprule
             & \multicolumn{3}{c}{Distance to Fire [km]} & \multicolumn{2}{c}{Speed [m/s]}  & \multicolumn{2}{c}{Comp. time [ms]} \\
& Minimum & Mean & Std. & Mean & Std. & Median & IQR\\
\midrule
Target & $\geq 0$ & 0.100 & - & 15.0 & - & - & - \\
\midrule
Nominal Planner    & -0.032     & 0.098    & 0.032    & 15.14                 & 0.73                & 27.32               & 4.37  & Unsafe              \\
FASTER~\cite{tordesillas2021faster}             & 0.040      & 0.101    & 0.030    & 12.60                 & 2.08                & 78.50               & 20.64            & Safe, but gets trapped in pocket   \\
Backup Filters~\cite{singletary2022onboard} & 0.081      & 0.240    & 0.054    & 10.11                 & 3.52                & 0.87$^*$                & 0.05       & Safe, but conservative and slow        \\
Gatekeeper (proposed)         & 0.049      & 0.108    & 0.034    & 14.91                 & 1.35                & 3.39                & 0.11          & Safe      \\ \bottomrule
\end{tabular}
      
      \label{tab:results}
      \vspace{-5mm}
  \end{table*}

We simulate an autonomous helicopter performing the firewatch mission, around a fire with an initial perimeter of 16~km. The helicopter starts 0.45~km from the fire, and is tasked to fly along the perimeter at a target airspeed of 15~m/s without entering the fire. The helicopter is modeled as
\begin{subequations}
\eqnN{
&\dot x_1 = x_3 \cos x_4  &\dot x_2 = x_3 \sin x_4\\
&\dot x_3 = u_1 &\dot x_4 = (g / x_3) \tan u_2,
}
\end{subequations}
where $x_1, x_2$ are the cartesian position coordinates of the helicopter wrt an inertial frame, $x_3$ is the speed of the vehicle along its heading, $x_4$ is the heading, and $g$ is the acceleration due to gravity. The control inputs are $u_1$, the acceleration along the heading, and $u_2$, the roll angle. The inputs are bounded, with $|u_1| < 0.5g$ and $|u_2| < \pi/4$~rad. This system models a UAV that can control its forward airspeed and makes coordinated turns. Notice the model has a singularity at $x_3 = 0$, and the system is \emph{not} control affine.

The fire is modeled using level-set methods~\cite{alessandri2021parameter}. In particular, the fire is described using the implicit function $\phi : \R \times \R^2 \to \R$, where $\phi(t, p)$ is the signed distance to the firefront from location $p$ at time $t$. Hence, the safe set is
\eqn{
\Scal(t) = \{ x : \phi(t, [x_1, x_2]^T) \geq 0 \} \notag
}
where $[x_1, x_2]$ are the Cartesian coordinates of the UAV.

The evolution of the fire is based on the Rothermel 1972 model~\cite{rothermel1972mathematical}. Given the \ac{RoS} function $\sigma: \R^2 \to \R$, the safe set evolves according to
\eqn{
  \frac{\partial \phi}{\partial t}(t, p) + \sigma(p) \norm { \nabla \phi(t, p) } = 0 \; \forall p \in \R^2
}
The \ac{RoS} depends on various environmental factors including terrain topology, vegetation type, and wind~\cite{rothermel1972mathematical, andrews2018rothermel} but can be bounded~\cite{cruz201910}. The simulated environment used a \ac{RoS} function that the controllers did not have access to. The only information the controllers could use was the thermal image (to detect the fire within a $\pm 1$~km range of the UAV) and the assumption that the maximum rate of spread is 8~km/h.

We compare our approach against the nominal planner and two state of the art methods for similar problems,~Fig.~\ref{fig:fire}. In particular, we compare (A) a nominal planner (black), (B) FASTER~\cite{tordesillas2021faster} (purple), (C) Backup Filters~\cite{singletary2022onboard} (blue) and (D)~\gatekeeper{} (green). Since these methods were not originally developed for dynamic environments with limited sensing, both methods (B, C) were modified to be applicable to this scenario. See~\cite{gatekeeperRepo} for details. Method (A) represents the baseline planner without any safety filtering. Methods (B), (C) and (D) are safety filtering methods that use the nominal planner of (A) and modify it to ensure safety.

The simulation environment and each of the methods were implemented in \texttt{julia}, to allow for direct comparison, using \texttt{Tsit5()}~\cite{rackauckas2017differentialequations} with default tolerances. Each run simulates a flight time of 50~minutes. The tracking controller was implemented as zero-order hold, updated at 20~Hz. Measurements of the firefront were available at 0.1~Hz, triggering the planners to update, intentionally slow to highlight the challenges of slow perception/planning systems. The measurements are a bitmask image, defining the domain where $\phi \leq 0$, at a grid resolution of 10~meters. These simulations were performed on a 2019 Macbook Pro (Intel i9, 2.3 GHz, 16 GB). 

In the nominal planner, a linear \ac{MPC} problem is solved to generate trajectories that fly along the local tangent 0.1~km away from firefront at 15~m/s. The planner uses a simplified dynamic model, a discrete-time double integrator. This convex quadratic program (QP) is solved using \texttt{gurobi}. The median computation time is 27~ms, using $N=40$ waypoints and a planning horizon of 120~seconds. The tracking controller is a nonlinear feedback controller based on differential flatness~\cite{agrawal2021constructive, martin2003flat}. When tracking nominal trajectories, the system becomes unsafe, going as far as $32$~m into the fire. 

In FASTER, the same double integrator model is assumed, and a similar \ac{MPC} problem is solved. We impose additional safety constraints, that the committed trajectory must lie within a safe flight corridor~\cite{liu2017planning} based on the signed distance field to the fire, corrected based on the maximum fire spread rate. While this approach does keep the helicopter outside the fire, it gets surrounded by the fire (Fig.~\ref{fig:fire}a). This is ultimately due to the fact that FASTER only plans trajectories over a finite planning horizon, and is therefore unable to guarantee recursive feasibility in a dynamic environment. Due to the large number of additional constraints on the QP, FASTER is about 3~times slower than the nominal planner.

In the Backup Filters approach, the backup trajectory is numerically forward propagated on the nonlinear system over the same 120~second horizon, and can be computed efficiently, requiring less than 1~ms per iteration.  Although this keeps the system safe, it does so at the cost of performance: the mean distance to the fire is 0.24~km, more than twice the target value, and the average speed is 10~m/s, 33\% less than the target. This is because the desired flight direction is perpendicular to the backup flight direction, and therefore the executed trajectory is always off-nominal. 

In \gatekeeper{}, the committed trajectories are constructed by maximizing the interval that the nominal trajectory is tracked, before implementing the backup controller. This allows the system to follow the nominal, and deviate only when required to ensure safety. As before, the nominal trajectory is 120~s long, and the backup is simulated for $T_B = 120$~s.  To initialize \gatekeeper{}, the first candidate trajectory is constructed using just the backup controller, effectively setting $T_s=0$. In our experiments this was sufficient ensure an initial valid committed trajectory.

In Fig.~\ref{fig:fire}c, we see that \gatekeeper{} chooses to not fly into the pocket, since it cannot ensure a safe path out of the pocket exists. \gatekeeper{} is computationally lightweight, with a median run time of 3.4~ms, more than 20~times faster than FASTER. This is because \gatekeeper{} searches over a scalar variable in a bounded interval, instead of optimizing $\R^{4N + 2N -2}$ variables as in the \ac{MPC} problem. 

We studied the effect of conservatism in the environment model. For instance, suppose we assumed the max fire spread rate was 16~km/hr instead of 8~km/hr. Simulations showed that \gatekeeper{} still maintains safe, but the resulting trajectories are more conservative: the mean distance to the fire increases by 37\% to 0.148~km, and the mean speed decreases by 19\% to 12.14~m/s. Despite doubling the level of conservatism in the environment model, we see a modest impact on the conservativeness of the resulting trajectories.

\subsection{Quadrotor Navigation (Simulations)}
\label{sec:quad_sim}

\begin{figure*}[t]
    \centering
    \includegraphics[width=\linewidth]{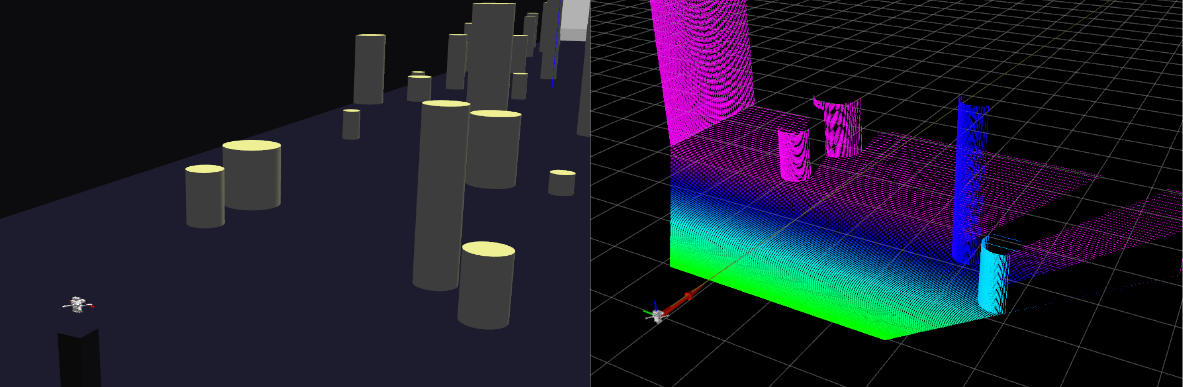}
    \caption{(Left) Simulation environment comprising of a quadrotor navigating in a 50~m long corridor with randomly scattered cylindrical obstacles of various heights and radii. This picture depicts the ``Easy 1" world. (Right) The point-cloud sensor data received by the quadrotor describing the environment. Using the point-cloud, a \ac{SDF} representation of the environment is constructed. A \ac{SFC}, i.e., a convex polyhedron of obstacle-free space,  centered on the quadrotor is extracted and used as the perceived safe set. The nominal planner treats unknown regions as free, while \gatekeeper{} treats unknown regions as occupied. }
    \label{fig:gazebo}
\end{figure*}
\begin{figure*}[t]
    \centering
    \includegraphics[width=\linewidth]{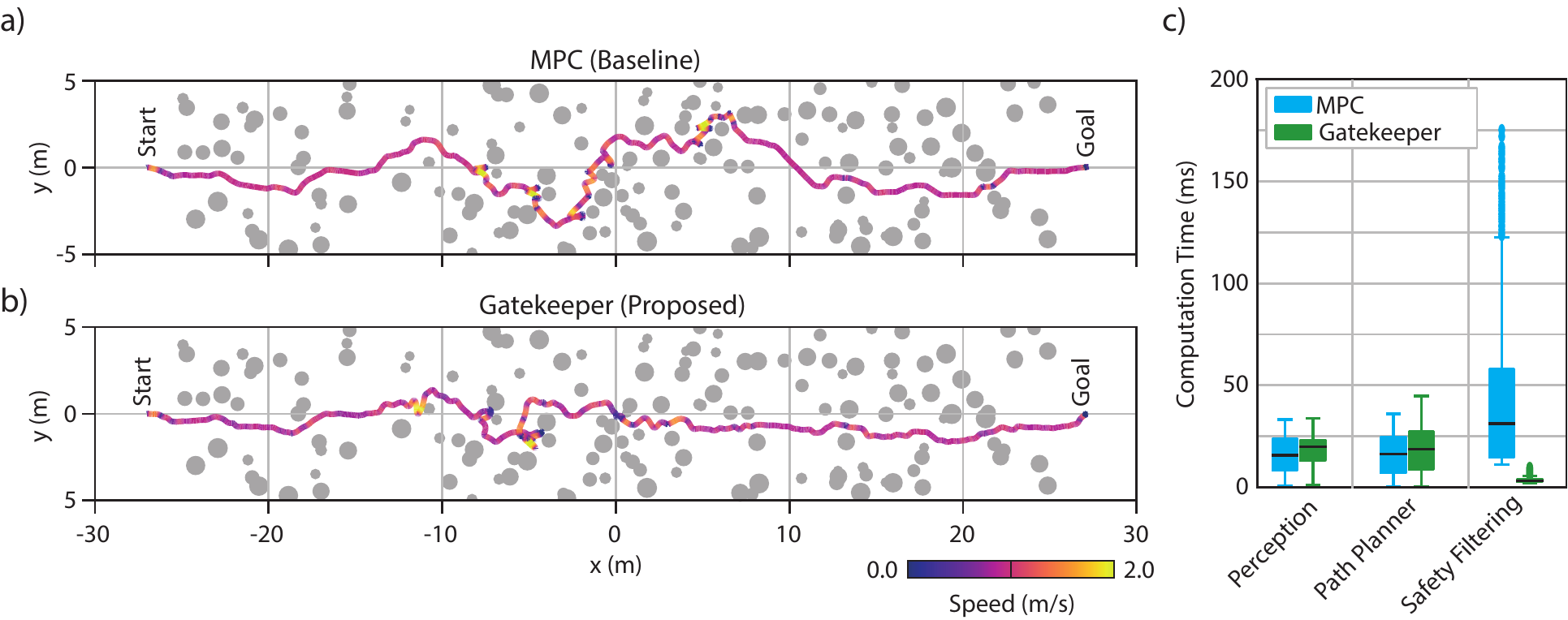}
    \caption{Trajectories executed in the ``Hard 1" world, using (a)~\ac{MPC}-based safety filter (baseline) and (b)~the proposed \gatekeeper{}-based safety filter. The gray circles indicate obstalces. Visually, the paths are similar, and are traversed with similar speeds. The color indicates that for most of the trajectories, the speed is at the target of 1~m/s, but near the obstacles (where there is greater replanning), the speeds vary more. (c) Box-and-whiskers plot showing the computation time for perception, planning, and safety filtering. The computation time for perception and path planning is similar with both safety filters, since both use the same perception and path planning implementation. However, \gatekeeper{} is significantly faster than the \ac{MPC}-based safety filter.}
    \label{fig:gazebo-paths}
\end{figure*}
\begin{table*}[t]
    \centering
    \caption{Summary of simulations in 15 different worlds with 3 difficulty levels, comparing the performance of an \ac{MPC}-based safety filter against \gatekeeper{}. \gatekeeper{} is able to successfully reach the goal in more scenarios, and is an order of magnitude computationally faster.}
\begin{tabular}{l rr rr rr rr}
\toprule
& \multicolumn{2}{c}{Goal Reached?} & \multicolumn{2}{c}{Median Comp. Time [ms]} & \multicolumn{2}{c}{Max Comp. Time [ms]} & \multicolumn{2}{c}{Average Speed [m/s]}\\
\cmidrule(lr){2-3} \cmidrule(lr){4-5} \cmidrule(lr){6-7} \cmidrule(l){8-9}
World & \multicolumn{1}{c}{\ac{MPC}} & \multicolumn{1}{c}{GK} & \multicolumn{1}{c}{\ac{MPC}} & \multicolumn{1}{c}{GK} & \multicolumn{1}{c}{\ac{MPC}} & \multicolumn{1}{c}{GK} & \multicolumn{1}{c}{\ac{MPC}} & \multicolumn{1}{c}{GK} \\
\midrule
easy 1 & True & True & 34.71 $\pm$ 0.10 & 3.28 $\pm$ 0.04 & 168.89 & 10.78 & 0.91 & 0.81 \\
easy 2 & True & True & 35.55 $\pm$ 0.11 & 3.42 $\pm$ 0.05 & 161.59 & 10.51 & 0.82 & 0.77 \\
easy 3 & True & True & 33.18 $\pm$ 0.12 & 3.26 $\pm$ 0.05 & 151.12 & 12.48 & 0.83 & 0.67 \\
easy 4 & True & True & 34.17 $\pm$ 0.25 & 3.33 $\pm$ 0.05 & 172.15 & 11.70 & 0.83 & 0.45 \\
easy 5 & True & True & 35.40 $\pm$ 0.20 & 3.36 $\pm$ 0.05 & 180.94 & 11.94 & 0.90 & 0.35 \\
medium 1 & True & True & 39.78   $\pm$ 0.20 & 3.27 $\pm$  0.06 & 178.21 & 10.62 & 0.78 & 0.61 \\
medium 2 & \color{red}{False} & True & 47.62  $\pm$ 1.10 & 3.27 $\pm$  0.05 & 217.97 & 12.19 & 0.57 & 0.76 \\
medium 3 & True & True & 33.65   $\pm$ 0.08 & 3.18 $\pm$  0.04 & 199.93 & 11.53 & 0.79 & 0.75 \\
medium 4 & \color{red}{False} & \color{red}{False} & 44.93 $\pm$ 2.03 & 3.23 $\pm$  0.06 & 199.74 & 9.25 & 0.44 & 0.43 \\
medium 5 & True & True & 27.72   $\pm$ 0.10 & 3.23 $\pm$  0.05 & 211.08 & 10.82 & 0.81 & 0.82 \\
hard 1 & True & True & 31.22     $\pm$ 0.16 & 3.18 $\pm$  0.07 & 201.54 & 9.65 & 0.68 & 0.82 \\
hard 2 & \color{red}{False} & True & 56.61    $\pm$ 0.61 & 3.41 $\pm$  0.08 & 184.23 & 12.06 & 0.68 & 0.79 \\
hard 3 & \color{red}{False} & \color{red}{False} & 44.75   $\pm$ 0.54 & 3.35 $\pm$  0.06 & 213.06 & 9.43 & 0.34 & 0.54 \\
hard 4 & True & True & 13.60     $\pm$ 0.09 & 3.25 $\pm$  0.04 & 218.98 & 10.41 & 0.34 & 0.73 \\
hard 5 & \color{red}{False} & True & 56.57    $\pm$ 4.26 & 3.25 $\pm$  0.06 & 208.30 & 10.15 & 0.50 & 0.68 \\
\bottomrule
\end{tabular}
    \label{tab:gk_sims}
\end{table*}

We demonstrate the efficacy of the \gatekeeper{} algorithm for a quadrotor flying through a previously unobserved area, in both a high fidelity simulation, and hardware experiments. The desired goal location is specified by the human operator. The quadrotor must simultaneously sense the environment, build a local map of the obstacles, plan a path to the goal, filter the path using \gatekeeper{}, and finally execute the committed trajectory. All of the processing happens onboard, in realtime, and by using \gatekeeper{}, the quadrotor does not crash into any obstacles. Each step of the perception-planning-control stack is described next, followed by a comparison to state-of-the-art methods. All simulations were run using Gazebo and RotorS~\cite{Furrer2016}, on an AMD Ryzen 7 5800h CPU 16~GB with a NVIDIA RTX 3050Ti. All hardware experiments were performed on a 16~GB Nvidia Xavier NX.

An environment with a forest of cylinders of random sizes, locations, and heights is generated in a corridor 50~m long, and 10~m wide. The start and goal locations are free, but a safe trajectory between these may not exist in environments with many obstacles. Since DMP is complete, the quadrotor will continue to explore until it finds a path to the goal.  

\subsubsection*{Perception} The quadrotor is equipped with a front-facing Intel Realsense D455 camera, which has a limited field of view of $87^\circ \times 58^\circ$, and a limited sensing range of $8$~meters, operating at 30~FPS. The incoming depth maps are fused into a \ac{ESDF} representation using the NvBlox package~\cite{nvblox} at a resolution of 7.5~cm. 

\subsubsection*{Path Planner}
From the ESDF, a 2D slice of the obstacle geometry between 0.8-1.2~m height is extracted. 
The path to the goal location is planned using the \ac{DMP}~\cite{liu2017planning}, a computationally efficient alternative to A*  which also pushes the path away from the obstacles. Unknown cells are treated as free cells. The planner takes less than 30~ms to replan trajectories, and is operated at 5~Hz. Given a desired linear travel speed~$v=2$~m/s, time is allocated to each leg of the returned path to construct the trajectory. This trajectory is not dynamically feasible for the quadrotor's nonlinear dynamics. 

\subsubsection*{Safety Filtering} In our implementation of \gatekeeper{}, a $4 \times 4 \times 2$~meter block centered on the quadrotor is extracted from the \ac{ESDF}. unknown voxels are treated as obstacles. A convex polyhedron representing the safe region is constructed using the DecompUtil~\cite{liu2017planning}, and is the safe set $\Bcal_k$ used in \gatekeeper{}. The environment is assumed static, but is unknown at the start of the run - as new regions are observed, the perceived safe set expands to include new regions. 

Next, \gatekeeper{} (as described in Algorithm~\ref{alg:full}) is used to convert the nominal trajectory into a dynamically feasible and safe trajectory for the quadrotor to follow.\footnote{We used a triple integrator model to validate trajectories, as in~\cite{lopez2017aggressive, tordesillas2021faster, tal2020accurate}. We tried the nonlinear model in~\cite{lee2010geometric}, but the communication latency between the Pix32 and Xavier NX degraded performance.} The backup controller used is a stopping controller. For $p^{can}$ to be valid, it must (A) lie within the safe polyhedron mentioned above (accounting for the grid resolution (0.075~m), the quadrotor's radius (0.15~m), and a robustness margin $R=0.1$~m)\footnote{Robustness margins were determined by flying the quadrotor, and measuring the tracking error as it executed some trajectories. The measured error was 5-10~cm, and thus $R=0.1$~m was chosen.}, (B) terminate within the safe polyhedron accounting for the quadrotor radius and a robustness margin $R+r = 0.2$~m, (C) terminate with zero speed and zero control input. These conditions guarantee that the quadrotor can hover indefinitely at the terminal position.\footnote{Here, we do not consider the quadrotor's limited battery life as a constraint. This is addressed in~\cite{naveed2023eclares} using the \gatekeeper{} strategy.} The maximum switch time was $T_S \leq 2$~s, and the backup trajectory was propagated from $T_b = 2.0$~s. Note, due to the safety filtering, the nominal trajectory may not be traversable, for instance through a narrow passage. When this occurs, \gatekeeper{} publishes a virtual obstacle, forcing the nominal planner to replan alternative routes. 

\subsubsection*{Tracking Controller} The last committed trajectory tracked using a geometric tracking controller~\cite{lee2010geometric}, running at 250~Hz.

\subsubsection*{Benchmark} Our implementation of \gatekeeper{} is compared with an \ac{MPC} safety filter. In the \ac{MPC} filter, the following optimization problem is solved:
\eqnN{
\argmin{x \in \Xcal^{N+1}, u \in \Ucal^{N}} & \sum_{i = 0}^{N} \norm{x_i - [p^{nom}_k]_i}_Q^2 +  \sum_{i=0}^{N-1}\norm{u_i - [u^{nom}_k]_i}_R^2\\
\text{s.t. } & x_{i+1} = Ax_i + Bu_i\\
& x_i \in \Bcal_k\\
& x_0 = \hat x(t_k), \quad x_{N} = x_{N-1}
}
where $[p^{nom}_k]_i = p^{nom}_k(t_k + i \Delta T)$ where $\Delta T = 0.02$~seconds is the discretization step size. A planning horizon of 2 seconds is considered, same as in our \gatekeeper{} implementation.  $[u^{nom}_k]_i$ the corresponding control input. To avoid solving a nonlinear program, the dynamics model assumed for the \ac{MPC} safety filter is the linear double integrator. The set $\Bcal_k$ is the same safe flight polyhedron used in \gatekeeper{}. The initial condition is required to match with the estimated state, and the last constraint ensures that the quadrotor trajectory terminates within the horizon. The resulting problem is a quadratic program, and solved using OSQP~\cite{osqp}.

\subsubsection*{Results}

Fifteen different world environments were constructed with 3 difficulty levels, defined by the density of obstacle cylinders (Fig.~\ref{fig:gazebo}). The quadrotors were tasked to fly a linear distance of 54~meters, at a desired speed of 1~m/s. Table~\ref{tab:gk_sims} summarizes the performance with the \ac{MPC} safety filter and using \gatekeeper{}. Both the \ac{MPC} and \gatekeeper{} algorithms prevented collisions. In some cases neither \ac{MPC} nor \gatekeeper{} were able to reach the goal location, although \gatekeeper{} was able to find a trajectory in more cases than \ac{MPC}. \ac{MPC} was consistently slower than the \gatekeeper{}, requiring approximately 10x the computation time. Finally the average speed of the quadrotor was similar with both filters. Through this, we conclude that the performance of \gatekeeper{} and \ac{MPC} are similar, although \gatekeeper{} computationally efficient, while additionally handling the nonlinear dynamics of the quadrotor.

\subsection{Quadrotor Navigation (Experiments)}
\label{sec:quad_exp}

\begin{figure}
    \centering
    \includegraphics[width=0.95\linewidth]{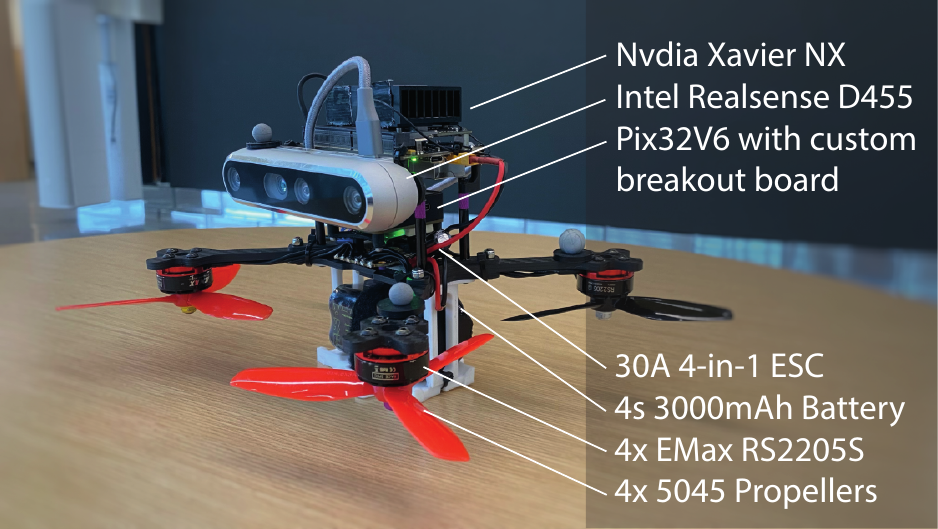}
    \caption{Quadrotor used for experiments. A combination of off-the-shelf components and custom breakout boards is used to minimize weight and maximize performance.}
    \label{fig:quad_components}
\end{figure}

\begin{figure*}
    \centering
    \includegraphics[width=0.98\linewidth]{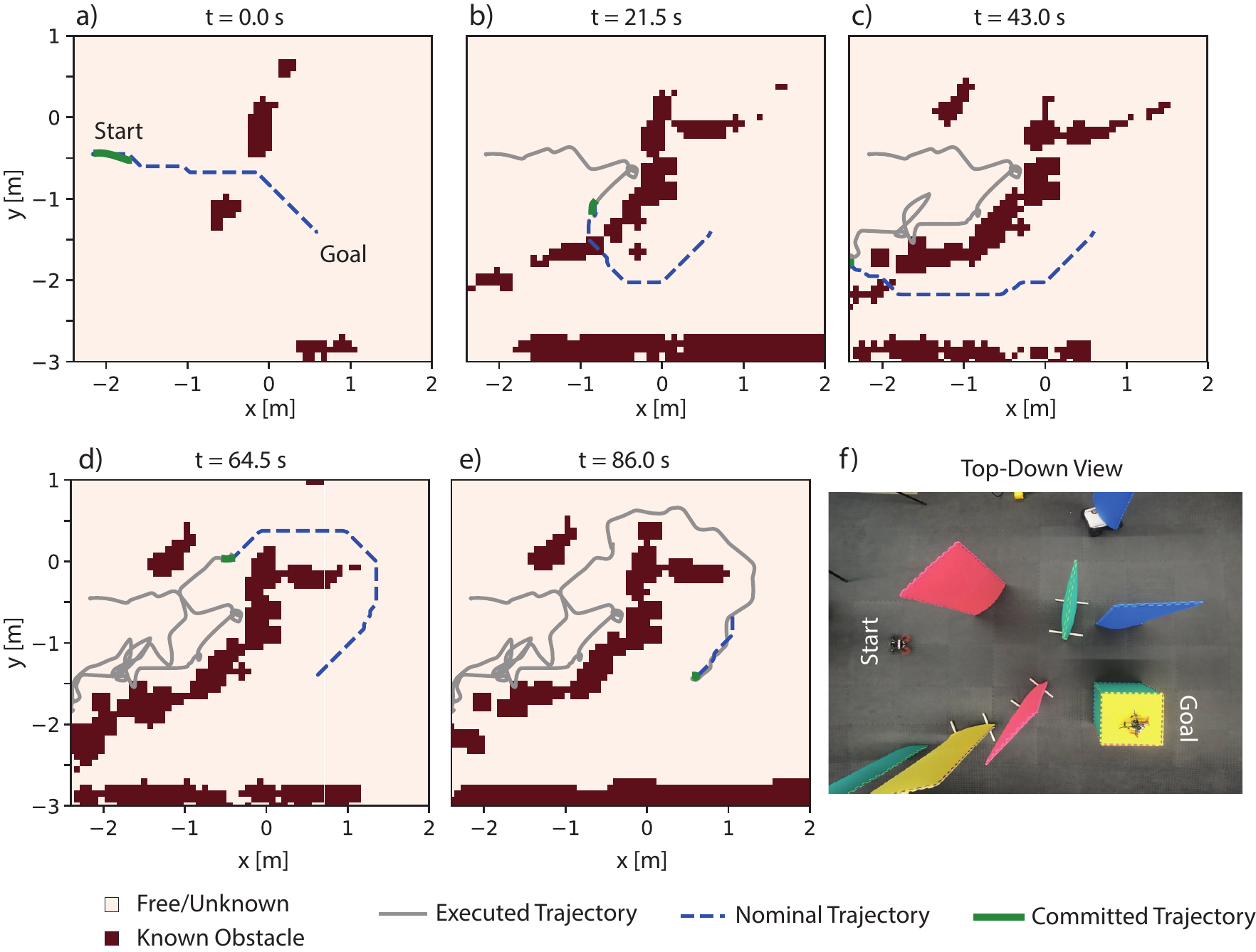}
    \caption{(a-e) Snapshots of the map, nominal trajectory, committed trajectory, and executed path of the quadrotor. (f) Top-down view of the obstacle geometry. }
    \label{fig:experiment-path}
\end{figure*}

\begin{figure}
    \centering
    \includegraphics[width=0.98\linewidth]{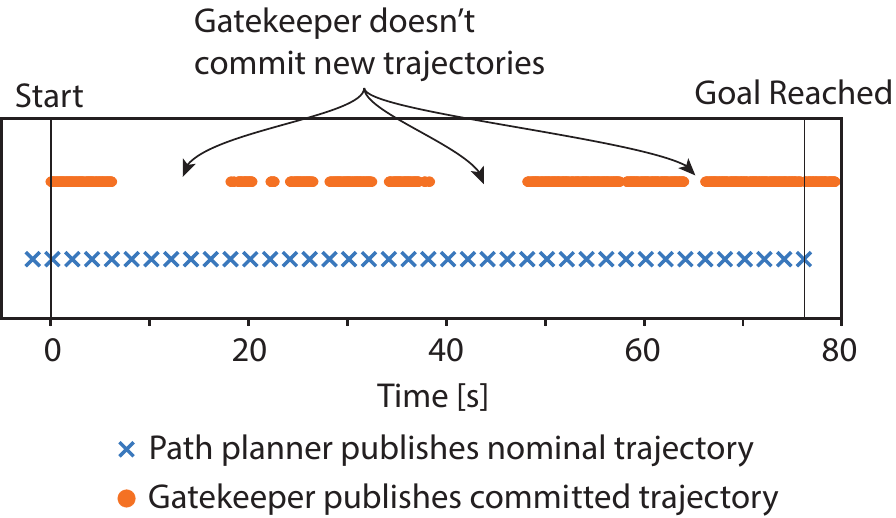}
    \caption{Each dot (cross) represents a timepoint when the committed (nominal) trajectory is published. The gaps represent intervals when \gatekeeper{} prevents unsafe trajectories from being committed. During these times, the controller continues to track the the last committed trajectory.}
    \label{fig:experiment-timings}
\end{figure}

We also demonstrate the algorithm experimentally. A custom quadrotor was designed to optimize the payload, and maximize the flight time (Fig.~\ref{fig:quad_components}). The quadrotor's wet weight is 820~g, with a 15~min hover flight time. The perception, planning, and safety filtering steps were all performed on an onboard computer, the NVIDIA Xavier NX. The low-level geometric controller~\cite{lee2010geometric} was implemented on a Pix32V6c, communicating with the Xavier over UART. The goal destination and yaw angle was specified by a human operator. 

As in the simulations, the quadrotor uses the Realsense D455 camera's RGBD images to construct a map of the environment using NvBlox, plans a trajectory using \ac{DMP}, and filters the trajectory using \gatekeeper{}.\footnote{In hardware, the path planner replans once every 2~seconds. The ESDF is updated at 5~Hz, and \gatekeeper{} is run at 20~Hz.} The last committed trajectory is tracked using the geometric controller. Each of these steps were implemented as described in Section~\ref{sec:quad_sim}.

Figure~\ref{fig:experiment-path} shows top-down snapshots of the map, and both the nominal and committed trajectories. See the video in~\cite{gatekeeperRepo}. Initially, the quadrotor tries to fly through a gap between the green and red obstacles. However, since the gap is too small for \gatekeeper{} to certify that it is safe to traverse through the gap,\footnote{The minimum gap required is the sum of (quadrotor diameter, 0.3~m) + (voxel size of map, 0.075~m) + (tracking radius $r$, 0.1~m) + (robustness radius $R$~0.1~m) = 0.58~m. The gap was measured to be 0.45~m across.} new trajectories are not committed, and the quadrotor executes its backup controller: stop and yaw. Once the nominal planner plans a new trajectory towards the right \gatekeeper{} allows a new trajectory to be committed. However, as the quadrotor approaches this gap, again it is too narrow to safely traverse. This repeats a few times before eventually the nominal planner finds the trajectory that indeed is safe to traverse, and the quadrotor reaches the goal destination. 

In Fig.~\ref{fig:experiment-timings}, the times at which nominal and committed trajectories are published are plotted. In our implementation using ROS2, when no new candidate trajectories are valid (as in~\eqref{eqn:dont_update}), the \gatekeeper{} node does not publish a new committed trajectory, and therefore the controller continues to track the last committed trajectory. Therefore, in Fig.~\ref{fig:experiment-timings}, the path-planner publishes at regular intervals, but there are gaps when \gatekeeper{} is running but not publishing new committed trajectories. To allow the system to continue making progress towards the goal, we publish a virtual obstacle along the nominal trajectory when this happens, forcing the path planner to find a new trajectory to the goal. In this particular run, we observed 9 such instances. 

The experiments also highlighted some limitations of \gatekeeper{} that can form the basis for future study. In particular, suppose a nominal planner is poorly designed, and produces trajectories that are collision free, but not desirable, e.g., if the nominal plan causes the drone to jerk back and forth and yaw rapidly. Such a nominal trajectory could pass the validity check~\ref{def:robustly_valid}, but could lead to the quadrotor chattering. In the future, we wish to investigate how to co-design the \gatekeeper{} with planners and controllers to avoid such undesired trajectories. In our experiments, we have sometimes observed the nominal planner making large and abrupt changes in the nominal trajectory, but the quadrotor was able to track the committed trajectories.

Finally, further investigation into the design of backup controllers could yield interesting directions for future research. In our current implementation, the backup controller stops the quadrotor along the nominal trajectory. However, the position at which the quadrotor comes to a stop could be designed, for example, to maximize the visibility of the unknown regions. Such a backup controller would still maintain safety but also allow the quadrotor to reason more efficiently about the environment it is operating in. Furthermore, to operate this quadrotor in an environment with dynamic obstacles further attention will be needed on the design of backup controllers. If one were to assume a bounded speed at which the environment could move, the safe flight polyhedrons could quickly collapse to becoming empty. Using semantic maps, for example~\cite{Rosinol20icra-Kimera}, might help to identify the dynamic parts of the environment, and overcome this issue.

\section{CONCLUSION}

This paper proposes an algorithm (``\gatekeeper{}") to safely control nonlinear robotic systems while information about dynamically-evolving safe states is received online. The algorithm constructs an infinite-horizon committed trajectory from a nominal trajectory using backup controllers. By extending a section of the nominal trajectory with the backup controller, \gatekeeper{} is able to follow nominal trajectories closely, while guaranteeing a safe control input is known at all times. We have implemented the algorithm in a simulated aerial firefighting mission and on-board a real quadrotor, where we demonstrated \gatekeeper{} is less conservative than similar methods, while remaining computationally lightweight.  Various comparisons to state-of-the-art techniques are also provided. 

A key benefit of the \gatekeeper{} approach is its applicability in dynamic environments where the safe set is sensed online. This allows the method to be applied to a wide range of scenarios where only limited safety information is known, for instance, overtaking and merging scenarios for autonomous vehicles. A limitation of \gatekeeper{} is the difficulty in finding backup controllers and sets that are suitable for the robotic system and environment considered, particularly in time-varying environments. Ultimately the possible safety guarantees rely on the ability to make forecasts of the environment given limited sensing information. Furthermore, when there are multiple safety conditions that must all be satisfied simultaneously, one could either design a single backup controller to satisfy all the constraints, or design multiple separate backup controllers and switch between them. Both of these approaches have their challenges and their suitability is case-dependent. These strategies require further analysis, an interesting direction for future work. Future directions also include developing more general methods to identify backup controllers, and understanding how the method can be applied in adversarial multi-agent settings.

\section*{Acknowledgements}

The authors would like to acknowledge Hardik Parwana and Kaleb Ben Naveed's assistance in the experiments.

\bibliographystyle{IEEEtran}
\bibliography{mybibfile}

\begin{thebibliography}{10}
\providecommand{\url}[1]{#1}
\csname url@rmstyle\endcsname
\providecommand{\newblock}{\relax}
\providecommand{\bibinfo}[2]{#2}
\providecommand\BIBentrySTDinterwordspacing{\spaceskip=0pt\relax}
\providecommand\BIBentryALTinterwordstretchfactor{4}
\providecommand\BIBentryALTinterwordspacing{\spaceskip=\fontdimen2\font plus
\BIBentryALTinterwordstretchfactor\fontdimen3\font minus
  \fontdimen4\font\relax}
\providecommand\BIBforeignlanguage[2]{{%
\expandafter\ifx\csname l@#1\endcsname\relax
\typeout{** WARNING: IEEEtran.bst: No hyphenation pattern has been}%
\typeout{** loaded for the language `#1'. Using the pattern for}%
\typeout{** the default language instead.}%
\else
\language=\csname l@#1\endcsname
\fi
#2}}

\bibitem{gatekeeperRepo}
D.~Agrawal and R.~Chen, ``Gatekeeper,''
  \url{https://github.com/dev10110/gatekeeper}, 2022.

\bibitem{lavalle2006planning}
S.~M. LaValle, \emph{Planning algorithms}.\hskip 1em plus 0.5em minus
  0.4em\relax Cambridge university press, 2006.

\bibitem{karaman2011anytime}
S.~Karaman, M.~R. Walter, A.~Perez, E.~Frazzoli, and S.~Teller, ``Anytime
  motion planning using the rrt,'' in \emph{2011 IEEE ICRA}.\hskip 1em plus
  0.5em minus 0.4em\relax IEEE, 2011, pp. 1478--1483.

\bibitem{webb2013kinodynamic}
D.~J. Webb and J.~Van Den~Berg, ``Kinodynamic \textrm{RRT}*: Asymptotically
  optimal motion planning for robots with linear dynamics,'' in \emph{2013 IEEE
  ICRA}, 2013, pp. 5054--5061.

\bibitem{richter2016polynomial}
C.~Richter, A.~Bry, and N.~Roy, ``Polynomial trajectory planning for aggressive
  quadrotor flight in dense indoor environments,'' in \emph{16th Int. Symposium
  on Robotics Research}.\hskip 1em plus 0.5em minus 0.4em\relax Springer, 2016,
  pp. 649--666.

\bibitem{tordesillas2021faster}
J.~Tordesillas, B.~T. Lopez, M.~Everett, and J.~P. How, ``Faster: Fast and safe
  trajectory planner for navigation in unknown environments,'' \emph{IEEE TRO},
  vol.~38, no.~2, pp. 922--938, 2021.

\bibitem{singletary2022onboard}
A.~Singletary, A.~Swann, Y.~Chen, and A.~D. Ames, ``Onboard safety guarantees
  for racing drones: High-speed geofencing with control barrier functions,''
  \emph{IEEE RAL}, vol.~7, no.~2, pp. 2897--2904, 2022.

\bibitem{gatekeeper_iros}
D.~R. Agrawal, R.~Chen, and D.~Panagou, ``Gatekeeper: Online safety
  verification and control for nonlinear systems in dynamic environments,'' in
  \emph{2023 IEEE IROS}, 2023.

\bibitem{harabor2011online}
D.~Harabor and A.~Grastien, ``Online graph pruning for pathfinding on grid
  maps,'' in \emph{Proceedings of the AAAI Conference on Artificial
  Intelligence}, vol.~25, no.~1, 2011, pp. 1114--1119.

\bibitem{liu2017planning}
S.~Liu, M.~Watterson, K.~Mohta, K.~Sun, S.~Bhattacharya, C.~J. Taylor, and
  V.~Kumar, ``Planning dynamically feasible trajectories for quadrotors using
  safe flight corridors in 3-d complex environments,'' \emph{IEEE RAL}, vol.~2,
  no.~3, pp. 1688--1695, 2017.

\bibitem{ames2019control}
A.~D. Ames, S.~Coogan, M.~Egerstedt, G.~Notomista, K.~Sreenath, and P.~Tabuada,
  ``Control barrier functions: Theory and applications,'' in \emph{2019 18th
  European Control Conference}, 2019, pp. 3420--3431.

\bibitem{cortez2020correct}
W.~S. Cortez and D.~V. Dimarogonas, ``Correct-by-design control barrier
  functions for euler-lagrange systems with input constraints,'' in \emph{2020
  American Control Conference}, 2020, pp. 950--955.

\bibitem{breeden2021guaranteed}
J.~Breeden and D.~Panagou, ``Guaranteed safe spacecraft docking with control
  barrier functions,'' \emph{IEEE Control Systems Letters}, vol.~6, pp.
  2000--2005, 2021.

\bibitem{abate2020enforcing}
M.~Abate and S.~Coogan, ``Enforcing safety at runtime for systems with
  disturbances,'' in \emph{2020 59th IEEE Conference on Decision and Control
  (CDC)}.\hskip 1em plus 0.5em minus 0.4em\relax IEEE, 2020, pp. 2038--2043.

\bibitem{llanes2021safety}
C.~Llanes, M.~Abate, and S.~Coogan, ``Safety from in-the-loop reachability for
  cyber-physical systems,'' in \emph{Proceedings of the Workshop on
  Computation-Aware Algorithmic Design for Cyber-Physical Systems}, 2021, pp.
  9--10.

\bibitem{agrawal2021constructive}
D.~R. Agrawal, H.~Parwana, R.~K. Cosner, U.~Rosolia, A.~D. Ames, and
  D.~Panagou, ``A constructive method for designing safe multirate controllers
  for differentially-flat systems,'' \emph{IEEE Control Systems Letters},
  vol.~6, pp. 2138--2143, 2021.

\bibitem{chen2021fastrack}
M.~Chen, S.~L. Herbert, H.~Hu, Y.~Pu, J.~F. Fisac, S.~Bansal, S.~Han, and C.~J.
  Tomlin, ``Fastrack: a modular framework for real-time motion planning and
  guaranteed safe tracking,'' \emph{IEEE Transactions on Automatic Control},
  vol.~66, no.~12, pp. 5861--5876, 2021.

\bibitem{bajcsy2019efficient}
A.~Bajcsy, S.~Bansal, E.~Bronstein, V.~Tolani, and C.~J. Tomlin, ``An efficient
  reachability-based framework for provably safe autonomous navigation in
  unknown environments,'' in \emph{2019 IEEE 58th Conference on Decision and
  Control (CDC)}.\hskip 1em plus 0.5em minus 0.4em\relax IEEE, 2019, pp.
  1758--1765.

\bibitem{bansal2017hamilton}
S.~Bansal, M.~Chen, S.~Herbert, and C.~J. Tomlin, ``Hamilton-jacobi
  reachability: A brief overview and recent advances,'' in \emph{2017 IEEE 56th
  Annual Conference on Decision and Control (CDC)}.\hskip 1em plus 0.5em minus
  0.4em\relax IEEE, 2017, pp. 2242--2253.

\bibitem{gurriet2019scalable}
T.~Gurriet, M.~Mote, A.~Singletary, E.~Feron, and A.~D. Ames, ``A scalable
  controlled set invariance framework with practical safety guarantees,'' in
  \emph{2019 IEEE CDC}.\hskip 1em plus 0.5em minus 0.4em\relax IEEE, 2019, pp.
  2046--2053.

\bibitem{choi2021robust}
J.~J. Choi, D.~Lee, K.~Sreenath, C.~J. Tomlin, and S.~L. Herbert, ``Robust
  control barrier--value functions for safety-critical control,'' in \emph{2021
  60th IEEE Conference on Decision and Control (CDC)}.\hskip 1em plus 0.5em
  minus 0.4em\relax IEEE, 2021, pp. 6814--6821.

\bibitem{tonkens2022refining}
S.~Tonkens and S.~Herbert, ``Refining control barrier functions through
  \textrm{H}amilton-\textrm{J}acobi reachability,'' in \emph{2022 IEEE/RSJ
  International Conference on Intelligent Robots and Systems}, 2022, pp.
  13\,355--13\,362.

\bibitem{dawson2022safe}
C.~Dawson, Z.~Qin, S.~Gao, and C.~Fan, ``Safe nonlinear control using robust
  neural lyapunov-barrier functions,'' in \emph{Conference on Robot
  Learning}.\hskip 1em plus 0.5em minus 0.4em\relax PMLR, 2022, pp. 1724--1735.

\bibitem{dawson2022learning}
C.~Dawson, B.~Lowenkamp, D.~Goff, and C.~Fan, ``Learning safe, generalizable
  perception-based hybrid control with certificates,'' \emph{IEEE Robotics and
  Automation Letters}, vol.~7, no.~2, pp. 1904--1911, 2022.

\bibitem{so2023train}
O.~So, Z.~Serlin, M.~Mann, J.~Gonzales, K.~Rutledge, N.~Roy, and C.~Fan, ``How
  to train your neural control barrier function: Learning safety filters for
  complex input-constrained systems,'' \emph{arXiv preprint arXiv:2310.15478},
  2023.

\bibitem{liu2023safe}
S.~Liu, C.~Liu, and J.~Dolan, ``Safe control under input limits with neural
  control barrier functions,'' in \emph{Conference on Robot Learning}.\hskip
  1em plus 0.5em minus 0.4em\relax PMLR, 2023, pp. 1970--1980.

\bibitem{lavanakul2024safety}
W.~Lavanakul, J.~J. Choi, K.~Sreenath, and C.~J. Tomlin, ``Safety filters for
  black-box dynamical systems by learning discriminating hyperplanes,''
  \emph{arXiv preprint arXiv:2402.05279}, 2024.

\bibitem{rawlings2017model}
J.~B. Rawlings, D.~Q. Mayne, and M.~Diehl, \emph{Model predictive control:
  theory, computation, and design}.\hskip 1em plus 0.5em minus 0.4em\relax Nob
  Hill Publishing Madison, WI, 2017, vol.~2.

\bibitem{rosolia2017autonomous}
U.~Rosolia, A.~Carvalho, and F.~Borrelli, ``Autonomous racing using learning
  model predictive control,'' in \emph{2017 American Control Conference
  (ACC)}.\hskip 1em plus 0.5em minus 0.4em\relax IEEE, 2017, pp. 5115--5120.

\bibitem{zhou2021raptor}
B.~Zhou, J.~Pan, F.~Gao, and S.~Shen, ``Raptor: Robust and perception-aware
  trajectory replanning for quadrotor fast flight,'' \emph{IEEE TRO}, vol.~37,
  no.~6, pp. 1992--2009, 2021.

\bibitem{chou2022safe}
G.~Chou, ``Safe end-to-end learning-based robot autonomy via integrated
  perception, planning, and control,'' Ph.D. dissertation, University of
  Michigan, Ann Arbor, 2022.

\bibitem{karkus2019differentiable}
P.~Karkus, X.~Ma, D.~Hsu, L.~P. Kaelbling, W.~S. Lee, and T.~Lozano-P{\'e}rez,
  ``Differentiable algorithm networks for composable robot learning,''
  \emph{arXiv preprint arXiv:1905.11602}, 2019.

\bibitem{kousik2020bridging}
S.~Kousik, S.~Vaskov, F.~Bu, M.~Johnson-Roberson, and R.~Vasudevan, ``Bridging
  the gap between safety and real-time performance in receding-horizon
  trajectory design for mobile robots,'' \emph{The International Journal of
  Robotics Research}, vol.~39, no.~12, pp. 1419--1469, 2020.

\bibitem{rothermel1972mathematical}
R.~C. Rothermel, \emph{A mathematical model for predicting fire spread in
  wildland fuels}.\hskip 1em plus 0.5em minus 0.4em\relax Intermountain Forest
  \& Range Experiment Station, Forest Service, US~…, 1972, vol. 115.

\bibitem{andrews2018rothermel}
P.~L. Andrews, ``The rothermel surface fire spread model and associated
  developments: A comprehensive explanation,'' \emph{Gen. Tech. Rep.
  RMRS-GTR-371. Fort Collins, CO: US Dept. of Agriculture, Forest Service,
  Rocky Mountain Research Station. 121 p.}, vol. 371, 2018.

\bibitem{khalil2002nonlinear}
H.~K. Khalil, ``Nonlinear systems, 3rd edition,'' \emph{Prentice Hall}, 2002.

\bibitem{agrawal2022safe}
D.~R. Agrawal and D.~Panagou, ``Safe and robust observer-controller synthesis
  using control barrier functions,'' \emph{IEEE Control Systems Letters},
  vol.~7, pp. 127--132, 2022.

\bibitem{blanchini1999set}
F.~Blanchini, ``Set invariance in control,'' \emph{Automatica}, vol.~35,
  no.~11, pp. 1747--1767, 1999.

\bibitem{gurriet2018towards}
T.~Gurriet, A.~Singletary, J.~Reher, L.~Ciarletta, E.~Feron, and A.~Ames,
  ``Towards a framework for realizable safety critical control through active
  set invariance,'' in \emph{2018 ACM/IEEE 9th International Conference on
  Cyber-Physical Systems (ICCPS)}, 2018, pp. 98--106.

\bibitem{tordesillas2021mader}
J.~Tordesillas and J.~P. How, ``Mader: Trajectory planner in multiagent and
  dynamic environments,'' \emph{IEEE TRO}, vol.~38, no.~1, pp. 463--476, 2021.

\bibitem{oleynikova2017voxblox}
H.~Oleynikova, Z.~Taylor, M.~Fehr, R.~Siegwart, and J.~Nieto, ``Voxblox:
  Incremental 3d euclidean signed distance fields for on-board mav planning,''
  in \emph{2017 IEEE/RSJ IROS}.\hskip 1em plus 0.5em minus 0.4em\relax IEEE,
  2017, pp. 1366--1373.

\bibitem{agrawal2024online}
D.~R. Agrawal, R.~Govindjee, J.~Yu, A.~Ravikumar, and D.~Panagou, ``Online and
  certifiably correct visual odometry and mapping,'' \emph{arXiv preprint
  arXiv:2402.05254}, 2024.

\bibitem{kanayama1990stable}
Y.~Kanayama, Y.~Kimura, F.~Miyazaki, and T.~Noguchi, ``A stable tracking
  control method for an autonomous mobile robot,'' in \emph{IEEE ICRA}.\hskip
  1em plus 0.5em minus 0.4em\relax IEEE, 1990, pp. 384--389.

\bibitem{lee2010geometric}
T.~Lee, M.~Leok, and N.~H. McClamroch, ``Geometric tracking control of a
  quadrotor uav on se (3),'' in \emph{49th IEEE CDC}.\hskip 1em plus 0.5em
  minus 0.4em\relax IEEE, 2010, pp. 5420--5425.

\bibitem{martin2003flat}
P.~Martin, R.~M. Murray, and P.~Rouchon, ``Flat systems, equivalence and
  trajectory generation,'' \emph{CDS Technical Report}, 2003.

\bibitem{chen2021backup}
Y.~Chen, M.~Jankovic, M.~Santillo, and A.~D. Ames, ``Backup control barrier
  functions: Formulation and comparative study,'' in \emph{2021 60th IEEE
  Conference on Decision and Control}, 2021, pp. 6835--6841.

\bibitem{gurriet2019realizable}
T.~Gurriet, P.~Nilsson, A.~Singletary, and A.~D. Ames, ``Realizable set
  invariance conditions for cyber-physical systems,'' in \emph{2019 IEEE
  ACC}.\hskip 1em plus 0.5em minus 0.4em\relax IEEE, 2019, pp. 3642--3649.

\bibitem{rackauckas2017differentialequations}
C.~Rackauckas and Q.~Nie, ``\texttt{Differentialequations.jl}--a performant and
  feature-rich ecosystem for solving differential equations in
  \texttt{julia},'' \emph{J. Open Research Software}, vol.~5, no.~1, 2017.

\bibitem{alessandri2021parameter}
A.~Alessandri, P.~Bagnerini, M.~Gaggero, and L.~Mantelli, ``Parameter
  estimation of fire propagation models using level set methods,''
  \emph{Applied Mathematical Modelling}, vol.~92, pp. 731--747, 2021.

\bibitem{cruz201910}
M.~G. Cruz and M.~E. Alexander, ``The 10\% wind speed rule of thumb for
  estimating a wildfire’s forward rate of spread in forests and shrublands,''
  \emph{Annals of Forest Science}, vol.~76, no.~2, pp. 1--11, 2019.

\bibitem{Furrer2016}
\BIBentryALTinterwordspacing
F.~Furrer, M.~Burri, M.~Achtelik, and R.~Siegwart, \emph{Robot Operating System
  (ROS): The Complete Reference (Volume 1)}.\hskip 1em plus 0.5em minus
  0.4em\relax Cham: Springer International Publishing, 2016, ch. RotorS---A
  Modular Gazebo MAV Simulator Framework, pp. 595--625. [Online]. Available:
  \url{http://dx.doi.org/10.1007/978-3-319-26054-9_23}
\BIBentrySTDinterwordspacing

\bibitem{nvblox}
\BIBentryALTinterwordspacing
{NVIDIA}, ``{NvBlox}.'' [Online]. Available:
  \url{https://github.com/nvidia-isaac/nvblox}
\BIBentrySTDinterwordspacing

\bibitem{lopez2017aggressive}
B.~T. Lopez and J.~P. How, ``Aggressive collision avoidance with limited
  field-of-view sensing,'' in \emph{2017 IEEE/RSJ International Conference on
  Intelligent Robots and Systems}, 2017, pp. 1358--1365.

\bibitem{tal2020accurate}
E.~Tal and S.~Karaman, ``Accurate tracking of aggressive quadrotor trajectories
  using incremental nonlinear dynamic inversion and differential flatness,''
  \emph{IEEE Transactions on Control Systems Technology}, vol.~29, no.~3, pp.
  1203--1218, 2020.

\bibitem{naveed2023eclares}
K.~B. Naveed, D.~Agrawal, C.~Vermillion, and D.~Panagou, ``Eclares:
  Energy-aware clarity-driven ergodic search,'' \emph{arXiv preprint
  arXiv:2310.06933}, 2023.

\bibitem{osqp}
\BIBentryALTinterwordspacing
B.~Stellato, G.~Banjac, P.~Goulart, A.~Bemporad, and S.~Boyd, ``{OSQP}: an
  operator splitting solver for quadratic programs,'' \emph{Mathematical
  Programming Computation}, vol.~12, no.~4, pp. 637--672, 2020. [Online].
  Available: \url{https://doi.org/10.1007/s12532-020-00179-2}
\BIBentrySTDinterwordspacing

\bibitem{Rosinol20icra-Kimera}
\BIBentryALTinterwordspacing
A.~Rosinol, M.~Abate, Y.~Chang, and L.~Carlone, ``Kimera: an open-source
  library for real-time metric-semantic localization and mapping,'' in
  \emph{IEEE Intl. Conf. on Robotics and Automation (ICRA)}, 2020. [Online].
  Available: \url{https://github.com/MIT-SPARK/Kimera}
\BIBentrySTDinterwordspacing

\end{thebibliography}

\appendix
\subsection{Worked Example for the Firewatch Scenario}

This example demonstrates how the sets $\Scal(t), \Bcal_k(t), \Ccal_k(t)$ are related, using the firewatch mission. For simplicity, consider a double integrator system, 
\eqn{
\dot x &= Ax + B u
}
where $x_{pos} = [x_1, x_2]^T$ is the position of the helicopter, and $x_{vel} = [x_3, x_4]^T$ is the velocity. 

Say the fire starts at $t = t_0$, at location $p=[0,0]^T$. The fire expands radially, with rate of spread  $\sigma: \R^2 \to \Rnonneg$, i.e., $\sigma(p)$ is the rate of spread at a location $p \in \R^2$. To simplify the algebra, assume the \ac{RoS} depends only on $\norm{p}$, i.e, $\sigma(p_1) = \sigma(p_2)$ for any $\norm{p_1} = \norm{p_2}$. This means that the fire always spreads out uniformly in a circle. 

Therefore, the safe set is time-varying, described by
\eqn{
\Scal(t) = \left\{ x : \norm{x_{pos}} \geq \int_0^t \sigma(r(\tau)) d\tau \right\} \ \forall t \geq 0
}
where $r(t)$ is the radius of the fire at time $t\geq t_0$. 

Since we don't know $\sigma$, we don't know $\Scal(t)$. Instead, we assume a reasonable upper bound: $\sigma(r) \leq 2$~m/s for all $r\geq 0$. 

Therefore, at $t=t_0$, we can define an \emph{perceived safe set}:
\eqn{
\Bcal^0(t) = \{ x : \norm{x_{pos}} \geq 2 (t-t_0) \} \; \forall t \geq t_0
}
and clearly $\Bcal^0(t) \subset \Scal(t) \; \forall t \geq 0$. Notice that $\Bcal^0(t)$ is \emph{not} a controlled invariant set for the double integrator.\footnote{Technically, a higher-order \ac{CBF} could be used to design a QP controller that renders a subset of $\Bcal^0(t)$ forward invariant, but this is only possible since $\Bcal^0(t)$ is a sufficiently smooth function that we can analyze analytically.} 

Suppose the system can directly measure the fire's radius. Let the $k$-th measurement be $r_k = r(t_k)$. This allows us to define the $k$-th \emph{perceived} safe set:
\eqn{
\Bcal_k(t) = \{ x : \norm{x_{pos}} \geq r_k + 2 (t-t_k) \}  \; \forall t \geq t_k
}

One can verify
\eqnN{
\Bcal_k(t) &= \left\{ x : \norm{x_{pos}} \geq r_k + 2 (t-t_k) \right\}\\
&= \left\{ x : \norm{x_{pos}} \geq \int_{t_0}^{t_k} \sigma(r(\tau)) d\tau + 2 (t-t_k) \right\}\\
&\subset \left\{ x : \norm{x_{pos}} \geq \int_{t_0}^{t_k} \sigma(r(\tau)) d\tau + \int_{t_k}^{t} \sigma(r(\tau)) d\tau \right\}\\
&= \left\{ x : \norm{x_{pos}} \geq \int_{t_0}^{t} \sigma(r(\tau)) d\tau \right\}\\
& = \Scal(t)
}
i.e. $\Bcal_k(t) \subset \Scal(t)$ for all $t \geq t_k$.

Similarly, we can verify that for any $k \geq 0$, 
\eqnN{
\Bcal_{k+1}&(t) = \left\{ x : \norm{x_{pos}} \geq r_{k+1} + 2 (t-t_{k+1}) \right\}\\
&= \left\{ x : \norm{x_{pos}} \geq r_{k} + \int_{t_k}^{t_{k+1}} \sigma(r(\tau)) d\tau + 2 (t-t_{k+1}) \right\}\\
&\supset \left\{ x : \norm{x_{pos}} \geq r_{k} + 2(t_{k+1}  - t_{k}) + 2 (t-t_{k+1}) \right\}\\
&= \left\{ x : \norm{x_{pos}} \geq r_{k} + 2 (t-t_k) \right\}\\
&= \Bcal_{k}(t)
}
i.e., $\Bcal_k(t) \subset \Bcal_{k+1}(t)$ for all $t \geq t_k$.

This proves that Assumption~\ref{assump:bcal} is satisfied. Next,  we define the backup controllers.

For any $k \in \naturals$, suppose the state is $x_k = x(t_k)$. The backup controller should drive the system radially away from the fire. Define $n_k$ as the unit vector pointed at $x(t_k)$:
\eqn{
n_k = x_{pos}(t_k) / \norm{x_{pos}(t_k)}
}

Notice that if the position followed the reference
\eqn{
p_{ref}(t) = (1 + r_k + 2(t-t_k)) n_k
}
then the reference is moving radially at a speed of $2$~m/s, and therefore faster than the maximum spread rate of the fire. Thus $p_{ref}(t)$ is a safe trajectory for all $t \geq t_k$.

This leads to the following backup controller:
\eqn{
\pi_k^B(t, x) = - K \left (\bmat{x_{pos} \\ x_{vel} }  - \bmat{
p_{ref}(t)\\
2 n_k
} \right)
}
where $K \in \R^{2 \times 4}$ is a stabilizing LQR gain for the double integrator.  This controller stabilizes the system to $\Ccal_k(t)$, where
\eqn{
\Ccal_k(t) = \left\{ x : \norm{ x - \bmat{
p_{ref}(t)\\
2 n_k
}} \leq 1
\right\}
}
This set is controlled invariant using the backup controller $\pi_k^B$. Geometrically, $\Ccal_k(t)$ is a unit norm ball that is moving radially at 2~m/s in the $n_k$ direction. Therefore, $\Ccal_k(t) \subset \Scal(t)$ for all $t \geq t_k$, since the set is moving outwards radially at a speed higher than the maximum spread rate.

This example demonstrates how $\Scal(t)$, $\Bcal_k(t)$, $\Ccal_k(t)$ can be defined for a given problem. The main validation step in \gatekeeper{}, will confirm whether after following the nominal trajectory over $[t_k, t_k + T_S)$, the system is able to safely reach $\Ccal_k(t)$ using the backup controller $\pi_k^B$. 

While the sets were described analytically here, in simulations they were represented numerically using SDFs.

\begin{figure}[t]
    \centering
    \includegraphics[width=0.9\linewidth]{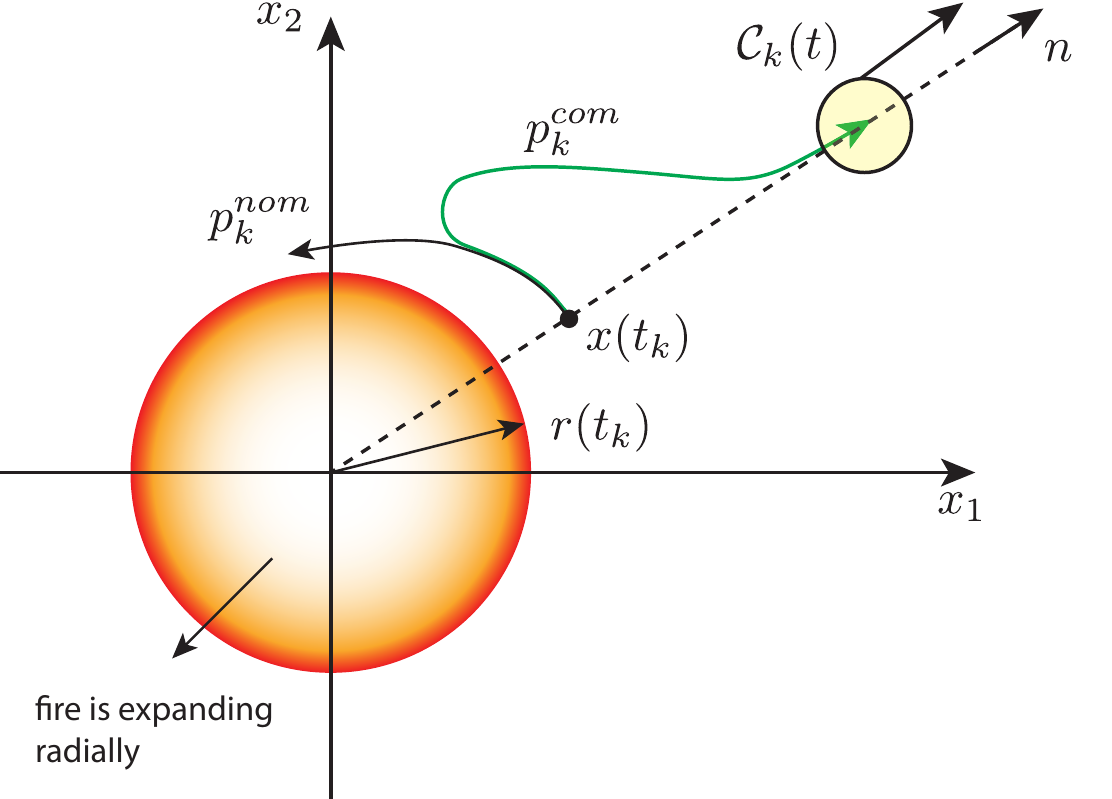}
    \caption{Depiction of the scenario in the worked example.}
    \label{fig:worked}
\end{figure}

\end{document}